\documentclass[11pt]{article}

\PassOptionsToPackage{hidelinks}{hyperref} 
\usepackage[final]{acl}
\usepackage{fvextra}
\usepackage{enumitem}
\usepackage{times}
\usepackage{latexsym}
\usepackage{multirow}
\usepackage{amsmath}
\usepackage{booktabs}
\usepackage{amssymb}
\usepackage{makecell}
\usepackage{subcaption}
\usepackage{listings}
\usepackage{xcolor}
\usepackage{colortbl} % 用于背景色或字体颜色
\usepackage{tabularx}
\usepackage{caption}
\usepackage{adjustbox}
\usepackage[most]{tcolorbox}
\tcbuselibrary{skins}
\tcbuselibrary{listings}
\tcbuselibrary{breakable}
\lstdefinestyle{jsonstyle}{
  basicstyle=\ttfamily\footnotesize,
  columns=fullflexible,
  keepspaces=true,
  showstringspaces=false,
  breaklines=true,
  breakatwhitespace=false,
  breakautoindent=false,
  breakindent=0pt,
  tabsize=2
}

\newtcolorbox{promptbox}[2][]{%
  enhanced,
  breakable,
  colback=white,
  colframe=black!65,
  boxrule=0.6pt,
  arc=1mm,
  boxsep=1mm,
  left=1.5mm,right=1.5mm,top=1mm,bottom=1mm,
  colbacktitle=black!80,
  coltitle=white,
  fonttitle=\bfseries\scriptsize,
  title={#2},
  before skip=4pt,
  after skip=4pt,
  #1
}

\definecolor{darkgreen}{RGB}{0, 100, 0}
\definecolor{darkred}{RGB}{139, 0, 0}
\usepackage[T1]{fontenc}
\usepackage[utf8]{inputenc}

\usepackage{microtype}

\usepackage{inconsolata}

\usepackage{graphicx}

\title{ExpertIVS: Sociological Expert Driven Individual Value Simulation in Large Language Models}

\author{
\textbf{Zhen Wang\textsuperscript{1,2}, Yuqi Ren\textsuperscript{2}\thanks{Corresponding authors.}, Yuehan Cui\textsuperscript{2}, Hongxiang Wang\textsuperscript{2}, Jianxiang Peng\textsuperscript{2},}\\
\textbf{Zhaoxia Zhang\textsuperscript{3}, Bingkun Zhu\textsuperscript{3}, Tongxuan Zhang\textsuperscript{4}, Dezhi Tong\textsuperscript{3}, Deyi Xiong\textsuperscript{2}\footnotemark[1]}\\
\textsuperscript{1}The International Joint Institute of Tianjin University, Fuzhou, Tianjin University, China\\
\textsuperscript{2}TJUNLP Lab, School of Computer Science and Technology, Tianjin University, China\\
\textsuperscript{3}National Governance Institute, Tianjin Normal University, China\\
\textsuperscript{4}College of Computer and Information Engineering, Tianjin Normal University, China\\
\texttt{\{tjwangzhen, ryq20, dyxiong\}@tju.edu.cn}
}

\begin{document}
\maketitle
\sloppy

% ===========================================================
% Abstract
% ===========================================================
\begin{abstract}
Large Language Model (LLM) agents have demonstrated considerable potential for social simulation, yet struggle to accurately model individual value systems. Most existing methods mechanically stitch survey responses into prompts, which suffer from semantic fragmentation, failing to capture the internal coherence of human value systems. The value systems of LLMs are typically assessed using static multiple-choice questions, which fail to evaluate the value orientation in real-world dialogue interactions. To address these issues, we propose \textbf{ExpertIVS}, a framework employing 14 Sociological Expert Agents to interpret World Values Survey (WVS) responses through structured professional perspectives, rather than direct responses concatenation. These expert agents perform deep semantic reconstruction to generate robust and internally consistent individual profiles. To evaluate the consistency between LLMs and individual value systems during dynamic interactions, we further introduce a multi-agent debate mechanism. Extensive experiments across 480 individuals from 12 countries demonstrate that ExpertIVS achieves 90.78\% value restoration fidelity and significantly outperforms baselines in value generalization (+5.3\%). Moreover, ExpertIVS exhibits strong personality discriminability and behavioral consistency, enabling a shift from mere response concatenation to genuine sociological role-playing.
\end{abstract}

\section{Introduction}

LLM-powered generative agents enable interactive simulations of human behaviors and social interactions \cite{park2023generative, peng2025diplomacyagent}, with extensions to social networks and society-scale simulations \cite{gao2023s3,piao2025agentsociety} and grounded individual modeling at scale \cite{park2024generative}. However, many existing simulations do not explicitly model individual value systems, which are essential for understanding why people behave differently under the same social context \cite{xu2024exploring,xu2025self}. Building value-grounded individuals from real-world data remains a key challenge \cite{mou2024individual, shi2024large}.

A common approach is to convert survey responses into natural language statements and use them as profiles, as exemplified by \textit{IndieValueCatalog} \cite{jiang2025can}. Other paradigms also condition agents with interviews \cite{park2024generative}, narrative backstories \cite{moon2024virtual}, or story-world grounding \cite{ran2025bookworld}, and scale persona resources for broader coverage \cite{ge2024scaling}. Despite progress, treating value profiling as flat fragments often fails to capture dependencies among value dimensions, weakening generalization beyond observed items \cite{jiang2025can}. Moreover, individual value profile is entangled with prompt sensitivity and potential bias risks. Sociodemographic personas can vary widely with prompt formulation \cite{lutz2025prompt} and may propagate stereotypes through linguistic signals \cite{sommerauer2025simulating}. Psychometric studies indicate that structured value frameworks and persona detail substantially affect individual simulated consistency \cite{jiang2023evaluating,zhu2024personality,bai2025scaling,li2025chatsop}. Therefore, equipping LLMs with a structured value system and capturing the intrinsic dependencies among values is crucial for simulating human individuals.

Evaluation of value consistency remains underexplored: many studies rely on static survey re-taking, but values are often reflected in dynamic deliberation and behavior under value conflict. Recent work highlights debate-dependent value dynamics \cite{sachdeva2025deliberative}, and both opportunities \cite{ki2025multiple} and risks \cite{taubenfeld2024systematic} of debate-based evaluation. Benchmarks on interactive social competence and cultural robustness further motivate dynamic, context-sensitive testing \cite{zhou2023sotopia,chiu2024culturalbench,kwok2024evaluating}.

In this paper, we propose \textbf{ExpertIVS}, a sociological expert driven individual value simulation framework. Specifically, we leverage top-tier LLMs to construct 14 expert agents, each corresponding to a specific value dimension in the WVS data 
\cite{haerpfer2022world}. From sociological perspectives, these expert agents conduct inductive reasoning of fragmented questionnaire responses, generating structured individual profiles with high generalization and logical coherence, thereby enhancing the realism of the simulated individuals. Additionally, to overcome the limitations of static evaluation, we design a multi-agent roundtable debate mechanism with three metrics (Value Alignment, Style Simulation and Persona Distinctiveness), that evaluates the value consistency between simulated individuals and real individuals under dynamic interactions by simulating high-density ideological debates. 

We comprehensively validated the effectiveness of our framework through experiments covering 480 real individuals across 12 different countries. First, we validated the value restoration capability of ExpertIVS through WVS questionnaire refilling. In the subsequent ``Leave-One-Out'' generalization tests, our framework achieved a 5.3\% improvement in accuracy over baseline in the results across value dimensions. This reveals that our framework not only accurately capture explicit value expressions, but also transform known value information into prior beliefs, providing critical inferential support. Additionally, we conducted extensive experiments to analyze the impact of demographic attributes on individual value simulation. Finally, dynamic evaluations based on the multi-agent debate show that our framework exhibits a high degree of consistency with real individuals in both positions and actions across multi-round debates. These findings verify the stability of our framework under complex interactions and provide a new paradigm for dynamically evaluating individual value simulation.

In a nutshell, our contributions are listed as follows:
\begin{enumerate}
    \item We propose a sociological expert-driven framework for individual value simulation, that constructs 14 sociological expert agents to summarize structured value systems.
    \item We design a multi-agent debate mechanism and three metrics (Value Alignment, Style Simulation and Persona Distinctiveness) to quantitatively assess the value consistency of simulated individuals in dynamic opinion interactions.
    \item Through extensive experiments, we demonstrate that our framework outperforms other methods in both value restoration fidelity and value generalization capability, and we further analyze the impact of demographic attributes on individual value simulation.
\end{enumerate}

\section{Related Work}
\label{sec:related_work}

\begin{figure*}[t!]
    \centering
    \includegraphics[width=\textwidth]{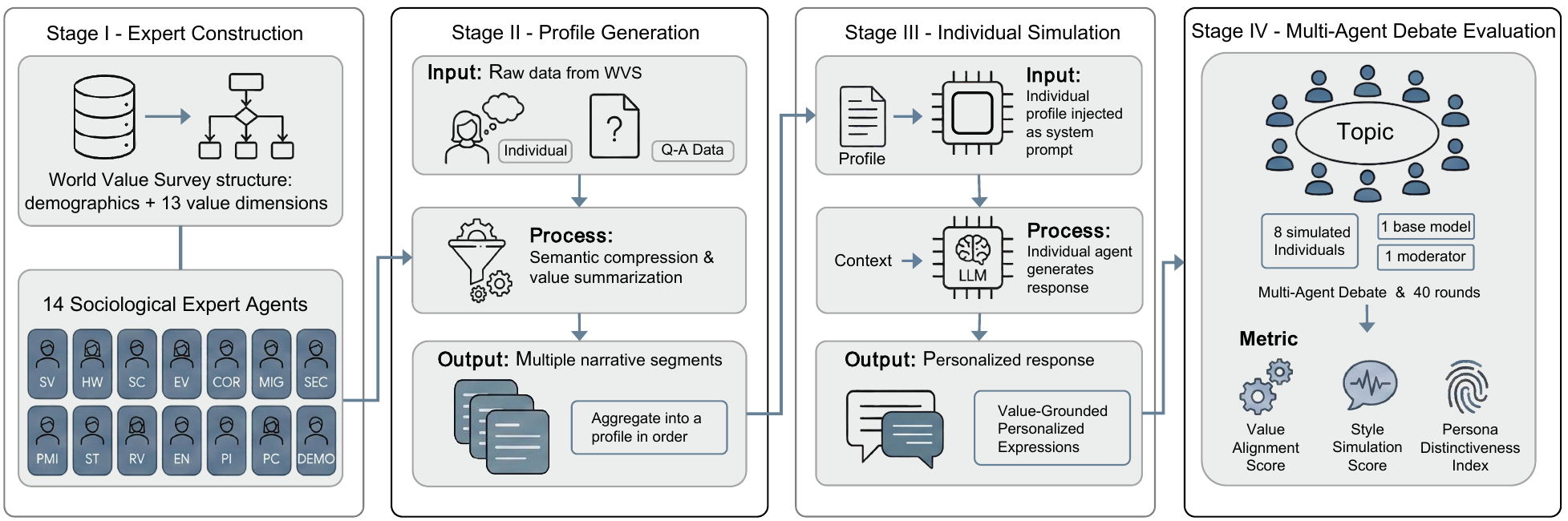}
    \caption{Diagram of the ExpertIVS framework. Following the WVS data structure, we constructed 14 sociologist expert agents and reconstructed the corresponding WVS data into individual value profiles. These individual value profiles are used as system prompts to simulate individuals. Next, the value consistency of individual simulation in dynamic interactions is evaluated by multi-agent debate.}
    \label{fig:framework}
\end{figure*}

LLM-driven simulations span across individual, scenario, and society \cite{mou2024individual}. We focus on individual value simulation and propose interaction-grounded evaluation for value consistency.

\subsection{Individual Simulation}

Early individual simulation largely relies on prompt-based role playing \cite{park2023generative}. Later work shows sociodemographic persona prompting is highly sensitive and can affect fidelity and stereotyping \cite{lutz2025prompt}, motivating more reproducible testbeds for pluralistic alignment and persona-based evaluation \cite{castricato2025persona}. Data-driven profiling conditions agents on richer evidence: WVS-based individualistic value reasoning \cite{jiang2025can}, interview-grounded agents representing 1,000+ real individuals \cite{park2024generative}, narrative backstories that condition LLMs on open-ended life stories to produce more representative, consistent, and diverse virtual personas for approximating human survey responses \cite{moon2024virtual}, and book-grounded agent societies \cite{ran2025bookworld}. Persona resources further scale coverage for synthesis and simulation \cite{ge2024scaling}.

A parallel line studies personality as a structured axis of individual differences, including psychometric evaluation and induction via MPI \cite{jiang2023evaluating}, personality alignment with large inventories \cite{zhu2024personality}, and evidence that persona detail drives simulation quality \cite{bai2025scaling}. Beyond simulation, personalized alignment methods optimize for diverse user preferences through training-time or inference-time control \cite{jang2023personalized,han2024value}, while raising questions about responsible bounds \cite{kirk2024benefits}.

However, persona-based simulation may propagate stereotypes \cite{sommerauer2025simulating} and trigger stereotype-driven implicit personalization \cite{neplenbroek2025reading}; profile generation can show stereotype and deviation biases \cite{wang2025measuring}. Evaluation is often static (survey re-taking), but value consistency requires behavioral, multi-turn testing: value--action gaps \cite{shen2025mind} and deliberation dynamics \cite{sachdeva2025deliberative} motivate interaction-grounded protocols. Debate can improve cultural parity \cite{ki2025multiple} yet may inherit systematic biases from base models \cite{taubenfeld2024systematic}. 

\subsection{Social Simulation}
Social simulation studies interaction dynamics and emergent macro phenomena. $S^3$ models diffusion in social networks \cite{gao2023s3}; Generative Agents demonstrate emergent behaviors in a sandbox town \cite{park2023generative}; \textit{Artificial Leviathan} explores order emergence through a social-contract lens \cite{dai2024artificial}; \textit{AgentSociety} scales to 10k+ agents for computational social experiments \cite{piao2025agentsociety}. For evaluation, \textit{Sotopia} provides interactive assessment of social intelligence in goal-driven interactions \cite{zhou2023sotopia}, while cultural benchmarks expose robustness gaps and cue sensitivity \cite{chiu2024culturalbench,kwok2024evaluating}. 
Despite these advances, most social simulations still rely on shallow persona descriptions or shared priors, which can lead to value homogenization and make emergent outcomes hard to interpret in terms of micro-level mechanisms \cite{mou2024individual,park2024generative}. Moreover, when simulations involve deliberation or negotiation, agents may exhibit drift across turns, revealing a mismatch between stated positions and interaction-grounded behaviors \cite{shen2025mind,sachdeva2025deliberative}. Few studies have quantitatively evaluated the robustness and consistency of deep-seated values within the context of high-intensity social deliberation.

\begin{table}[h] % 使用 [h] 让表格紧跟文字，或者 [t] 放在页顶
\centering
\small
\renewcommand{\arraystretch}{1.1} 
\begin{tabular}{cl}
\toprule
\textbf{ID} & \textbf{Dimension} \\
\midrule
1 & Social Values, Attitudes \& Stereotypes (SV)\\
2 & Happiness and Well-being (HW)\\
3 & Social Capital, Trust \& Membership (SC)\\
4 & Economic Values (EV)\\
5 & Corruption (COR)\\
6 & Migration  (MIG)\\
7 & Security  (SEC)\\
8 & Postmaterialist Index (PMI)\\
9 & Science \& Technology (ST)\\
10 & Religious Values (RV)\\
11 & Ethical Values and Norms (EN)\\
12 & Political Interest \& Political Participation (PI)\\
13 & Political Culture \& Political Regimes (PC)\\
14 & Demographics (Demo)\\
\bottomrule
\end{tabular}
\caption{The 14 sociological expert agents.}
\label{tab:experts}
\end{table}

\section{ExpertIVS}
\label{sec:methodology}

ExpertIVS is designed to enhance the individual value simulation by extracting the underlying value systems from survey responses. It comprises four sequential stages as Figure \ref{fig:framework}: (i) \textbf{Expert Construction}, creating 14 sociological expert agents; (ii) \textbf{Profile Generation}, yielding individual profiles for WVS respondents based on survey responses; (iii) \textbf{Individual Simulation}, instantiating individual agents through in-context learning; (iv) \textbf{Multi-Agent Debate Evaluation}, assessing the individual value simulation in dynamic interactions.

% ===========================================================
% Figure 5: Debate Framework (Double Column)
% Placed at the top of Section 4.5
% ===========================================================
\begin{figure*}[t]
    \centering
    \includegraphics[width=0.95\textwidth]{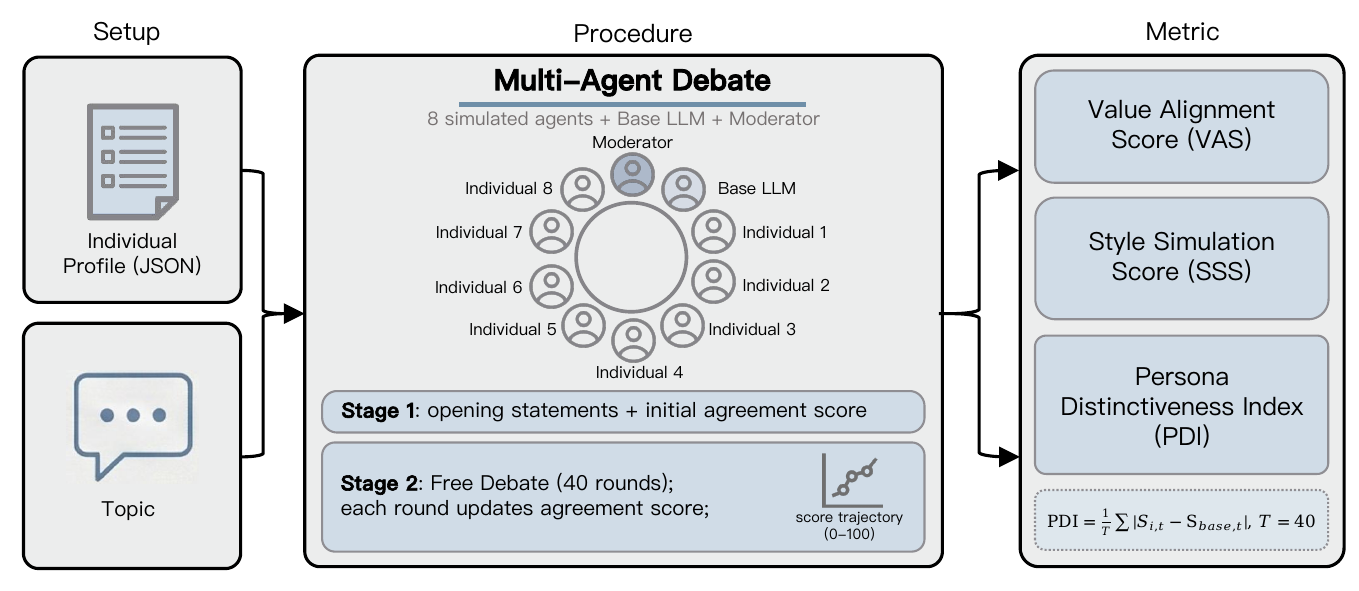} % 0.8 -> 更小/更大
    \caption{The Multi-Agent Debate workflow. Eight simulated agents and one Base LLM engage in a two-stage discussion (Opening Statement + 40 Rounds of Free Debate).}
    \label{fig:debate_framework}
\end{figure*}
% ===========================================================

\subsection{Expert Construction}
\label{subsec:data_structuring}

Following the WVS structure, which covers demographic attributes and 13 value dimensions, we constructed 14 independent sociological expert agents: one focusing on demographic information and the remaining 13 responsible for specific value dimensions. Each expert is assigned a dimension-specific system instruction responsible for summarizing questionnaire data within its designated dimension. Every expert prompt is shown as an example in Appendix \ref{app:expert_prompts}. All expert agents are created based on Gemini 2.5-flash. The complete list of experts is presented in Table \ref{tab:experts}.

\subsection{Profile Generation}
\label{subsec:pipeline}
The individual profile is composed of the specific value descriptions generated by the 14 expert agents. Specifically, given the raw WVS questionnaire and response of individual $i$ within value dimension $k$, the corresponding expert agent executes semantic compression and value summarization through in-context learning, generating a natural language narrative segment $s_{i,k}$. The specific in-context learning prompt is provided in Appendix \ref{app:expert_prompts}. This process differs from simple raw data concatenation. It requires the expert to distill logically consistent value beliefs from the data.

Then, the comprehensive profile $\mathcal{P}_i$ of individual $i$ is formed by the aggregation of outputs from all expert agents in accordance with the value dimensions of the WVS questionnaire:
\begin{equation}
\mathcal{P}_i = s_{i, \text{demo}} \oplus s_{i, \text{values}}
\end{equation}
where $s_{i, \text{demo}}$ provides demographic attributes, and $s_{i, \text{values}}$ represents the collection of the remaining 13 value segments. Together, they constitute the complete cognitive profile of the individual.

\subsection{Individual Simulation}
\label{subsec:instantiation}

For individual simulation, we inject the comprehensive profile $\mathcal{P}_i$ as the system prompt into the LLM to simulate the individual's value system. The detailed system prompt is provided in Appendix \ref{app:profile_example}.

Formally, given a context $C$, the response generation process of the individual agent, denoted as $r$, can be represented as a conditional probability distribution constrained by the individual profile $\mathcal{P}_i$:
\begin{equation}
P(r \mid C, \mathcal{P}_i) = \text{LLM}(C \mid \text{Prompt} = \mathcal{P}_i)
\end{equation}
By conditioning the LLM on profile $\mathcal{P}_i$, we guide the value orientation of LLM toward that of a specific individual, enabling personalized expressions grounded in the individual’s value logic and demographic background.

\subsection{Multi-Agent Debate Evaluation}
Individual simulation agents are commonly used in sociological simulation experiments, where they engage in dynamic interactions on open-ended questions. To evaluate the value consistency of individual agents in real-world scenarios, we designed a multi-agent debate mechanism. Specifically, eight individual simulation agents, along with a Base LLM (representing the LLM’s default value alignment), discuss a given topic, with a moderator (who does not contribute opinions on the topic) overseeing the overall process. The debate begins with each participant presenting their option statements, followed by 40 rounds of free debate. After each round, each agent updates its agreement score for the topic (ranging from 0 to 100) (see Figure~\ref{fig:debate_framework}). A detailed description of the multi-agent debate procedure is provided in Appendix \ref{app:nexus_procedure}.
\begin{table*}[t]
\centering
\small % Slightly smaller font to fit 15 columns
\setlength{\tabcolsep}{2.5pt} % Reduce column spacing
\begin{tabular}{lcccccccccccccc}
\toprule
\textbf{Country} & \textbf{AVG} & \textbf{COR} & \textbf{EV} & \textbf{EN} & \textbf{HW} & \textbf{MIG} & \textbf{PC} & \textbf{PI} & \textbf{PMI} & \textbf{RV} & \textbf{ST} & \textbf{SEC} & \textbf{SC} & \textbf{SV} \\
\midrule
Bolivia & 91.93  & 99.69  & 87.33  & 95.16  & 92.27  & 98.50  & 82.84  & 86.92  & 86.67  & 99.79  & 88.75  & 97.84  & 93.90  & 85.37   \\ 
Brazil & 90.03  & 98.13  & 84.58  & 91.30  & 92.96  & 94.00  & 80.50  & 79.58  & 91.67  & 98.33  & 78.75  & 97.14  & 91.94  & 91.50   \\ 
Britain & 90.89  & 97.50  & 78.33  & 91.48  & 94.55  & 97.50  & 82.29  & 82.08  & 92.50  & 97.00  & 87.92  & 96.67  & 95.38  & 88.44   \\ 
Canada & 91.72  & 100.00  & 74.58  & 85.87  & 95.68  & 99.75  & 82.09  & 87.78  & 95.00  & 96.67  & 88.75  & 100.00  & 93.93  & 92.33   \\ 
China & 93.56  & 99.13  & 80.83  & 92.28  & 97.73  & 97.19  & 86.46  & 96.41  & 93.33  & 98.33  & 95.42  & 99.75  & 92.69  & 86.75   \\ 
Egypt & 91.05  & 99.17  & 77.50  & 91.56  & 96.59  & 96.50  & 84.02  & 82.02  & 96.67  & 99.50  & 80.00  & 100.00  & 91.42  & 88.66   \\ 
Germany & 91.89  & 99.69  & 76.25  & 93.48  & 96.14  & 98.00  & 82.09  & 89.58  & 95.42  & 97.63  & 84.17  & 100.00  & 91.38  & 90.82   \\ 
India & 88.88  & 99.06  & 75.00  & 94.67  & 94.32  & 94.25  & 86.20  & 75.28  & 84.17  & 96.04  & 77.08  & 100.00  & 93.62  & 85.72   \\ 
Japan & 89.28  & 98.75  & 75.42  & 91.74  & 92.96  & 97.50  & 82.10  & 79.10  & 92.08  & 97.29  & 80.00  & 97.50  & 92.04  & 84.22   \\ 
Kenya & 90.91  & 100.00  & 85.42  & 90.10  & 91.30  & 99.75  & 80.83  & 83.50  & 86.08  & 99.58  & 88.67  & 97.86  & 92.62  & 86.19   \\ 
Russia & 88.08  & 95.63  & 68.33  & 90.76  & 94.55  & 97.25  & 82.20  & 77.43  & 91.25  & 97.29  & 77.92  & 97.50  & 90.85  & 84.11   \\ 
USA & 91.15  & 99.44  & 78.33  & 88.26  & 95.68  & 98.00  & 81.05  & 83.27  & 93.33  & 99.38  & 85.83  & 99.76  & 93.04  & 89.50   \\ 
\midrule
\textbf{Average} & 90.78 & 98.85 & 78.49 & 91.39 & 94.56 & 97.35 & 82.72 & 83.58 & 91.51 & 98.07 & 84.44 & 98.67 & 92.73 & 87.80 \\
\bottomrule
\end{tabular}
\caption{Results of value restoration fidelity across 12 countries and 13 value dimensions.}
\label{tab:voa_detailed}
\end{table*}
\paragraph{Evaluation Metrics}
We design three metrics to evaluate the alignment between the debate content and the individual’s value system:

\begin{itemize}
\item \textbf{Value Alignment Score (VAS).}
VAS measures whether the agent’s statements faithfully reflect the WVS responses and core values specified in the Profile.
\item \textbf{Style Simulation Score (SSS).}
SSS evaluates whether the agent’s linguistic style (e.g., tone and lexical complexity) aligns with the target demographic attributes.
\item\textbf{Persona Distinctiveness Index (PDI).}
PDI quantifies how much an agent’s scoring trajectory deviates from the Base LLM. Higher values indicate a more independent persona. For agent $i$, PDI is defined as:
\begin{equation}
\mathrm{PDI}_i = \frac{1}{T} \sum_{t=1}^{T} \left| S_{i,t} - S_{base,t} \right|
\end{equation}
where $S_{i,t}$ denotes agent $i$’s score at round $t$, $S_{base,t}$ denotes the Base LLM’s score.
\end{itemize}
Both VAS and SSS are scored by \textit{Gemini 3 Pro} and \textit{ChatGPT 5.2 Thinking}, and we report the average over five independent runs for robustness. The evaluation prompt is shown in Appendix \ref{app:restoration_eval_prompt}.

\section{Experiments}
\label{sec:experiments}
We systematically evaluated ExpertIVS by designing experiments that cover value restoration, value generalization capability, demographic ablations, and value consistency via multi-agent debate, as detailed below.
\subsection{Dataset and Setting}

%改成下面一段话
We sampled 480 real respondents from the WVS, spanning 12 countries across five continents, with 40 individuals per country (see Appendix \ref{app:country_individual_selection}). From each respondent’s WVS records, we constructed individual profiles by ExpertIVS, which are subsequently used in value restoration fidelity, generalization evaluations, demographic analysis, and multi-agent debate evaluation. Experiments were conducted using Gemini 2.5-flash and Deepseek V3.2 with default settings.

\subsection{Baseline and Metric}
We used the \textit{IndieValueCatalog} \cite{jiang2025can} as the baseline, which converts WVS questionnaires into natural language statements to represent an individual's value orientation (e.g., “I don’t believe in life after death”). In addition, we compared our method with two other profiling strategies: \textit{Simple Concatenation}, which directly concatenates the questionnaire questions and answers as the simulated individual's profile, and \textit{Anthology} \cite{moon2024virtual}, which transforms WVS questionnaires into an open-ended backstory for individual simulation. Comparisons with these two methods were conducted on 60 respondents (5 per country; the first five respondent IDs listed in Appendix~\ref{app:country_individual_selection}). Value Orientation Alignment (VOA) accuracy is used to evaluate the performance of individual agents when refilling the WVS questionnaire. VOA is an approximate accuracy metric for Likert-scale predictions. A prediction is considered correct if it matches the ground-truth tendency range (see Appendix~\ref{app:eval_metric} for the full rules).

\subsection{Value Restoration Fidelity}
\label{subsec:voa}

To evaluate the value restoration fidelity of ExpertIVS, we prompted the constructed individual agents to refill the blank WVS questionnaire. Using the VOA accuracy, we compared the responses generated by the individual agents with the ground-truth answers of real individuals. As shown in Table~\ref{tab:voa_additional}, both variants of ExpertIVS outperform \textit{Anthology}, and the performance remains strong despite a moderate drop when replacing Gemini 2.5-flash with DeepSeek V3.2.

\begin{table*}[t]
\centering
\small
\setlength{\tabcolsep}{3.8pt}
\resizebox{\textwidth}{!}{%
\begin{tabular}{lcccccccccccccc}
\toprule
\textbf{Method} & \textbf{AVG} & \textbf{COR} & \textbf{EV} & \textbf{EN} & \textbf{HW} & \textbf{MIG} & \textbf{PC} & \textbf{PI} & \textbf{PMI} & \textbf{RV} & \textbf{ST} & \textbf{SEC} & \textbf{SC} & \textbf{SV} \\
\midrule
Anthology & 69.20 & 66.73 & 50.72 & 70.91 & 84.70 & 58.53 & 60.77 & 68.04 & 49.22 & 86.69 & 65.83 & 94.28 & 67.53 & 75.60 \\
Ours (DeepSeek) & 87.78 & 97.36 & \textbf{79.44} & 90.19 & 87.70 & 96.92 & 77.95 & 80.56 & 84.61 & \textbf{98.19} & \textbf{85.00} & \textbf{99.84} & 85.43 & 77.90 \\
Ours (Gemini) & \textbf{90.78} & \textbf{98.85} & 78.49 & \textbf{91.39} & \textbf{94.56} & \textbf{97.35} & \textbf{82.72} & \textbf{83.58} & \textbf{91.51} & 98.07 & 84.44 & 98.67 & \textbf{92.73} & \textbf{87.80} \\
\bottomrule
\end{tabular}
}
\caption{Value restoration comparison with \textit{Anthology} across 13 dimensions. \textbf{Bold} indicates the best result.}
\label{tab:voa_additional}
\end{table*}

As shown in Tables~\ref{tab:voa_detailed} and \ref{tab:voa_additional}, ExpertIVS demonstrates strong value reproduction. In Tables~\ref{tab:voa_detailed}, across 12 countries, the average value restoration VOA accuracy reaches 90.78\%, indicating that the profiles generated by ExpertIVS enable the LLM to reliably reproduce core value orientations on most WVS questions. Specifically, the VOA accuracy exceeds 88\% for all countries, ranging from 88.08\% (Russia) to 93.56\% (China), which confirms strong cross-cultural universality. Moreover, cross-country differences are minimal, with a total population variance of 2.09 (SD = 1.45), suggesting that ExpertIVS maintains stable value fidelity when simulating individuals from diverse ideological and cultural systems, with no evident culture-specific bias.

Results across value dimensions indicate that strong fidelity in dimensions involving moral and fundamental beliefs. For instance, VOA accuracy in corruption (98.85\%), security (98.67\%), religious (98.07\%), and migration (97.35\%) are extremely high. These dimensions typically involve explicit identity labels or strong moral intuitions. This may arise from the fact that related questions in the WVS often contain clear semantic anchors, which are easy for experts to translate into unambiguous factual statements. In contrast, value fidelity in economy (78.49\%) and political cultural (82.72\%) dimensions is comparatively lower. These dimensions often involve complex competing value priorities. For example, economic issues frequently require subtle numerical positioning between efficiency and fairness. Individual profiles struggle to specify such fine-grained distinctions, resulting in fuzzy deviations.

\begin{table*}[t]
\centering
\resizebox{\textwidth}{!}{%
\begin{tabular}{l|cccccccccccccc}
\toprule
\textbf{Country} & \textbf{AVG} & \textbf{COR} & \textbf{EV} & \textbf{EN} & \textbf{HW} & \textbf{MIG} & \textbf{PC} & \textbf{PI} & \textbf{PMI} & \textbf{RV} & \textbf{ST} & \textbf{SEC} & \textbf{SC} & \textbf{SV}  \\
\midrule
Bolivia & 49.4/\textbf{50.9} & 41.9/\textbf{45.9} & 38.7/\textbf{41.2} & 66.7/\textbf{67.0} & 57.0/\textbf{57.7} & 44.2/\textbf{48.2} & 41.0/\textbf{41.1} & 43.8/\textbf{44.4} & \textbf{36.7}/33.8 & 65.2/\textbf{67.5} & 48.3/\textbf{52.5} & 52.9/\textbf{59.2} & \textbf{49.4}/44.3 & 56.9/\textbf{59.5} \\
Brazil & 47.0/\textbf{51.8} & 30.6/\textbf{37.5} & 35.0/\textbf{42.9} & 57.6/\textbf{59.6} & 49.5/\textbf{62.7} & 35.8/\textbf{36.2} & 42.2/\textbf{45.6} & 45.0/\textbf{50.2} & 39.6/\textbf{42.1} & 52.1/\textbf{62.7} & 45.0/\textbf{50.0} & 62.9/\textbf{65.2} & 56.4/\textbf{57.5} & 59.8/\textbf{61.0} \\
Britain & 54.1/\textbf{67.8} & 42.8/\textbf{66.6} & 32.1/\textbf{50.0} & 59.8/\textbf{71.1} & 73.2/\textbf{83.6} & 44.2/\textbf{63.2} & 55.6/\textbf{61.8} & 42.6/\textbf{59.1} & 35.8/\textbf{52.9} & 50.5/\textbf{73.0} & 60.0/\textbf{75.4} & 77.5/\textbf{81.3} & 62.9/\textbf{68.7} & 66.1/\textbf{74.1} \\
Canada & 56.5/\textbf{64.1} & 46.6/\textbf{53.4} & 39.2/\textbf{49.2} & 59.1/\textbf{65.2} & 74.5/\textbf{77.3} & 53.0/\textbf{67.5} & 59.3/\textbf{59.8} & 45.9/\textbf{57.2} & 32.9/\textbf{45.0} & 59.6/\textbf{73.5} & 65.0/\textbf{71.7} & 73.0/\textbf{76.5} & 61.5/\textbf{65.6} & 64.4/\textbf{70.8} \\
China & 57.6/\textbf{58.9} & 37.8/\textbf{43.9} & 35.0/\textbf{37.9} & 69.1/\textbf{70.7} & 65.9/\textbf{74.1} & 29.1/\textbf{35.9} & 49.5/\textbf{50.4} & \textbf{56.5}/46.7 & 43.8/\textbf{44.2} & 79.0/79.0 & 77.1/\textbf{79.2} & 68.5/\textbf{69.0} & 69.7/\textbf{69.9} & \textbf{67.6}/65.3 \\
Egypt & 58.2/\textbf{58.3} & \textbf{55.8}/40.8 & \textbf{54.6}/50.4 & 77.7/\textbf{79.1} & 52.3/\textbf{62.0} & 42.2/\textbf{45.8} & 41.2/\textbf{42.7} & \textbf{54.8}/54.4 & 43.3/\textbf{44.2} & \textbf{84.2}/83.2 & 43.8/\textbf{46.7} & 80.0/\textbf{81.0} & 54.4/\textbf{54.9} & 72.8/\textbf{73.1} \\
Germany & 58.1/\textbf{67.1} & 44.4/\textbf{58.3} & 37.9/\textbf{51.7} & 70.4/\textbf{75.5} & 79.3/\textbf{85.2} & 43.8/\textbf{58.0} & 61.1/\textbf{65.8} & 51.5/\textbf{65.9} & 37.1/\textbf{51.7} & 61.9/\textbf{71.0} & 59.2/\textbf{65.0} & 78.5/\textbf{82.6} & 63.7/\textbf{68.6} & 66.7/\textbf{72.6} \\
India & 48.1/\textbf{52.8} & 22.8/\textbf{35.0} & 37.9/\textbf{42.9} & 74.7/\textbf{75.1} & 55.0/\textbf{67.0} & 32.2/\textbf{34.8} & 45.6/\textbf{51.4} & 41.6/\textbf{48.5} & 32.9/\textbf{34.6} & 60.4/\textbf{64.4} & 48.7/\textbf{51.7} & 63.5/\textbf{65.7} & 46.4/\textbf{53.4} & \textbf{63.5}/62.5 \\
Japan & 49.8/\textbf{54.8} & 43.6/\textbf{53.9} & 33.3/\textbf{39.6} & 64.8/\textbf{67.0} & 65.9/\textbf{66.8} & 34.8/\textbf{47.2} & 49.7/\textbf{53.2} & 44.2/\textbf{48.6} & 38.8/\textbf{45.0} & 35.4/\textbf{42.9} & 57.1/\textbf{61.3} & 68.1/\textbf{69.8} & 56.4/\textbf{59.7} & 55.7/\textbf{57.6} \\
Kenya & 48.3/\textbf{49.3} & 40.3/\textbf{45.9} & 45.0/\textbf{50.0} & 55.3/\textbf{55.9} & 51.8/\textbf{52.3} & 39.5/\textbf{40.0} & 35.3/\textbf{35.5} & 42.2/\textbf{42.6} & \textbf{42.9}/37.1 & 70.4/\textbf{73.8} & 46.3/\textbf{47.9} & \textbf{56.0}/54.2 & 46.3/\textbf{47.7} & 56.3/\textbf{57.5} \\
Russia & 48.0/\textbf{50.9} & 27.5/\textbf{33.3} & 34.2/\textbf{38.3} & 54.5/\textbf{55.9} & 56.8/\textbf{67.3} & 41.5/\textbf{41.8} & 45.2/\textbf{47.2} & 40.6/\textbf{44.0} & 35.4/\textbf{36.3} & 48.3/\textbf{54.4} & \textbf{50.8}/\textbf{50.8} & 71.3/\textbf{73.2} & 57.0/\textbf{60.4} & \textbf{60.6}/58.3 \\
USA & 50.6/\textbf{61.7} & 39.7/\textbf{52.5} & 40.4/\textbf{61.2} & 54.3/\textbf{66.2} & 74.1/\textbf{77.0} & 37.8/\textbf{51.8} & 49.3/\textbf{58.0} & 41.4/\textbf{55.9} & 30.4/\textbf{39.2} & 44.2/\textbf{70.0} & 52.5/\textbf{63.8} & 73.6/\textbf{76.4} & 58.9/\textbf{62.8} & 61.1/\textbf{67.2} \\
\midrule
Average & 52.1/\textbf{57.4} & 39.5/\textbf{47.3} & 38.6/\textbf{46.3} & 63.7/\textbf{67.3} & 63.0/\textbf{69.4} & 39.8/\textbf{47.5} & 47.9/\textbf{51.0} & 45.8/\textbf{51.5} & 37.5/\textbf{42.2} & 59.3/\textbf{68.0} & 54.5/\textbf{59.7} & 68.8/\textbf{71.2} & 56.9/\textbf{59.5} & 62.6/\textbf{65.0} \\
\bottomrule
\end{tabular}%
}
\caption{Results of value generalization. The format is IndieValue / ExpertIVS. \textbf{Bold} indicates the higher performing method in that specific country-dimension cell. AVG is the mean over the 13 dimensions for each country.}
\label{tab:loo_results}
\end{table*}

\subsection{Value Generalization Capability}
\label{subsec:generalization}

To assess whether our framework captures the intrinsic relationships across value dimensions, rather than mechanically memorizing prompt inputs, we conducted a rigorous leave-one-out (LOO) generalization test, using Gemini 2.5-flash as the backbone model.

Specifically, for each value dimension, we masked the profile information related to that dimension in the system prompt, while retaining the other 12 value dimensions and demographic attributes. The individual agent then inferred questionnaire responses for the held-out dimension based on the profile generated by remaining values.

Results of value generalization are shown in Table \ref{tab:loo_results}. We observe that ExpertIVS achieves an average VOA accuracy of 57.4\%, representing a significant improvement of 5.3\% compared over the Baseline. We further compare ExpertIVS with \textit{Simple Concatenation} and \textit{Anthology}. As shown in Table~\ref{tab:loo_additional}, ExpertIVS achieves the best overall performance.

\begin{table*}[t]
\centering
\small
\setlength{\tabcolsep}{4.0pt}
\resizebox{\textwidth}{!}{%
\begin{tabular}{lcccccccccccccc}
\toprule
\textbf{Method} & \textbf{AVG} & \textbf{COR} & \textbf{EV} & \textbf{EN} & \textbf{HW} & \textbf{MIG} & \textbf{PC} & \textbf{PI} & \textbf{PMI} & \textbf{RV} & \textbf{ST} & \textbf{SEC} & \textbf{SC} & \textbf{SV} \\
\midrule
Simple Concatenation & 53.0 & 44.5 & 39.4 & 66.7 & 70.3 & 41.2 & 46.8 & 48.3 & 39.2 & 65.1 & 55.6 & 61.1 & 58.9 & 52.1 \\
Anthology & 54.5 & 46.9 & 40.3 & 66.2 & \textbf{71.1} & 41.9 & 47.0 & 46.2 & \textbf{42.7} & 67.6 & 55.8 & 60.4 & 58.7 & 63.4 \\
IndieValue & 52.1 & 39.5 & 38.6 & 63.7 & 63.0 & 39.8 & 47.9 & 45.8 & 37.5 & 59.3 & 54.5 & 68.8 & 56.9 & 62.6 \\
Ours & \textbf{57.4} & \textbf{47.3} & \textbf{46.3} & \textbf{67.3} & 69.4 & \textbf{47.5} & \textbf{51.0} & \textbf{51.5} & 42.2 & \textbf{68.0} & \textbf{59.7} & \textbf{71.2} & \textbf{59.5} & \textbf{65.0} \\
\bottomrule
\end{tabular}
}
\caption{Value generalization comparison with \textit{Simple Concatenation}, \textit{Anthology}, and \textit{IndieValue} across 13 dimensions. \textbf{Bold} indicates the best result.}
\label{tab:loo_additional}
\end{table*}

\begin{figure}[t]
    \centering
    \includegraphics[width= 1.0 \columnwidth]{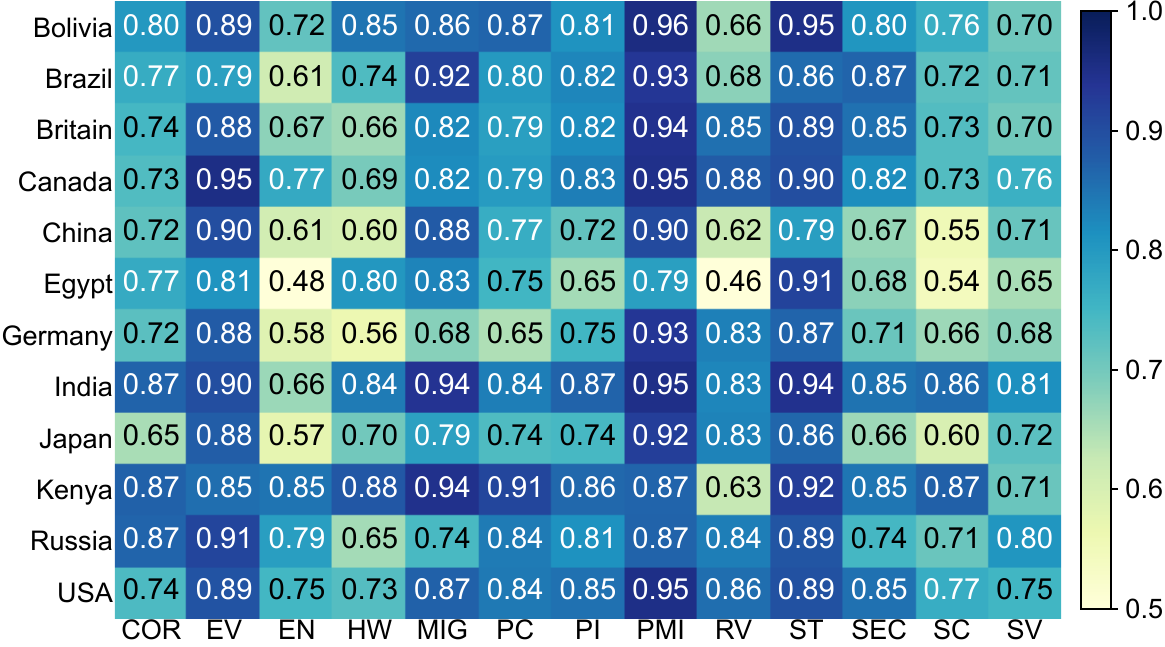}
    \caption{Heatmap of Shannon Entropy across different countries and value dimensions.}
    \label{fig:entropy_heatmap}
\end{figure}

To investigate the factors driving performance breaks down, we further employed Shannon entropy to measure the intrinsic uncertainty of each value dimension. From the Figure \ref{fig:entropy_heatmap}, we observe that VOA accuracy is strongly negatively correlated with entropy (Pearson $r=-0.58, p<0.001$). Values with low-entropy and broad consensus (e.g., Ethical Values) are easier to predict, while high-entropy, polarized value (e.g., Economic Values) are more difficult to predict.

% ===========================================================
% Placeholder for Entropy Heatmap Image
% ===========================================================
% ===========================================================

\paragraph{Cross-Country Analysis.} Country-level results confirm cross-cultural robustness of our framework. As shown, all 12 countries exhibit higher VOA accuracy, indicating consistent benefits from ExpertIVS are systematic rather than isolated gains in specific countries.

Countries with the most significant improvements include Britain (+13.7\%), USA (+11.1\%), and Germany (+9.0\%), whereas the improvements are relatively modest in countries such as Egypt (+0.1\%), Kenya (+1.0\%), and China (+1.3\%). We hypothesize that this disparity arises from differences in the internal structure of national value systems: countries with more pluralistic and internally conflicting value landscapes benefit more from structured value abstraction, as explicit value disentanglement helps resolve latent contradictions that are otherwise obscured by shallow data composition.

\paragraph{Cross-Value Analysis.} While high-entropy value dimensions are generally harder to predict, our method exhibits strong cross-value reasoning capabilities. As shown in Table~\ref{tab:loo_results}, the largest performance gains are achieved in Religious Values (+8.7\%), Corruption (+7.8\%), and Migration (+7.7\%). These value preferences are shaped by non-linear dependencies among multiple value cues and demographic factors (e.g., Corruption is difficult to predict from other surface-level raw data concatenation). This suggests that our framework is able to distill intrinsic associations among values from observed cues, enabling value consistent inference under missing information.

\subsection{Impact of Demographics}
\label{subsec:demographics}
Demographic attributes are widely regarded as antecedents of value formation in sociology \cite{miles2022demographic}. We quantify their contribution via an ablation study based on preceding experiments: (i) \textbf{Value Restoration} (re-taking the full WVS questionnaire) and (ii) \textbf{Value Generalization} (Leave-One-Out generalization test). Detailed experimental results are provided in the Appendix \ref{app:nodemo_results}.

From Figure \ref{fig:radar_repro}, we observe that removing demographics results in barely change across 12 countries, with the overall average VOA accuracy decreasing from 90.78\% to 90.54\%, consistent with findings in \textit{IndieValueCatalog} \cite{jiang2025can}. This indicates that expert-generated value narratives already provide dense, explicit signals, making demographics largely redundant when the value cues are directly available.
As shown in Figure~\ref{fig:radar_gen}, generalization is substantially affected across all countries, the overall average VOA accuracy drops from 57.4\% to 54.8\%. This supports when direct value evidence is unavailable, demographic attributes act as sociological priors \cite{fiske1990continuum}, guiding the LLM to infer plausible answers in uncertain situations.
% ===========================================================
% Figure 3: Radar Charts
% ===========================================================
\begin{figure}[t]
    \centering
    \begin{subfigure}[t]{0.48\columnwidth}
        \centering
        \includegraphics[width=\linewidth]{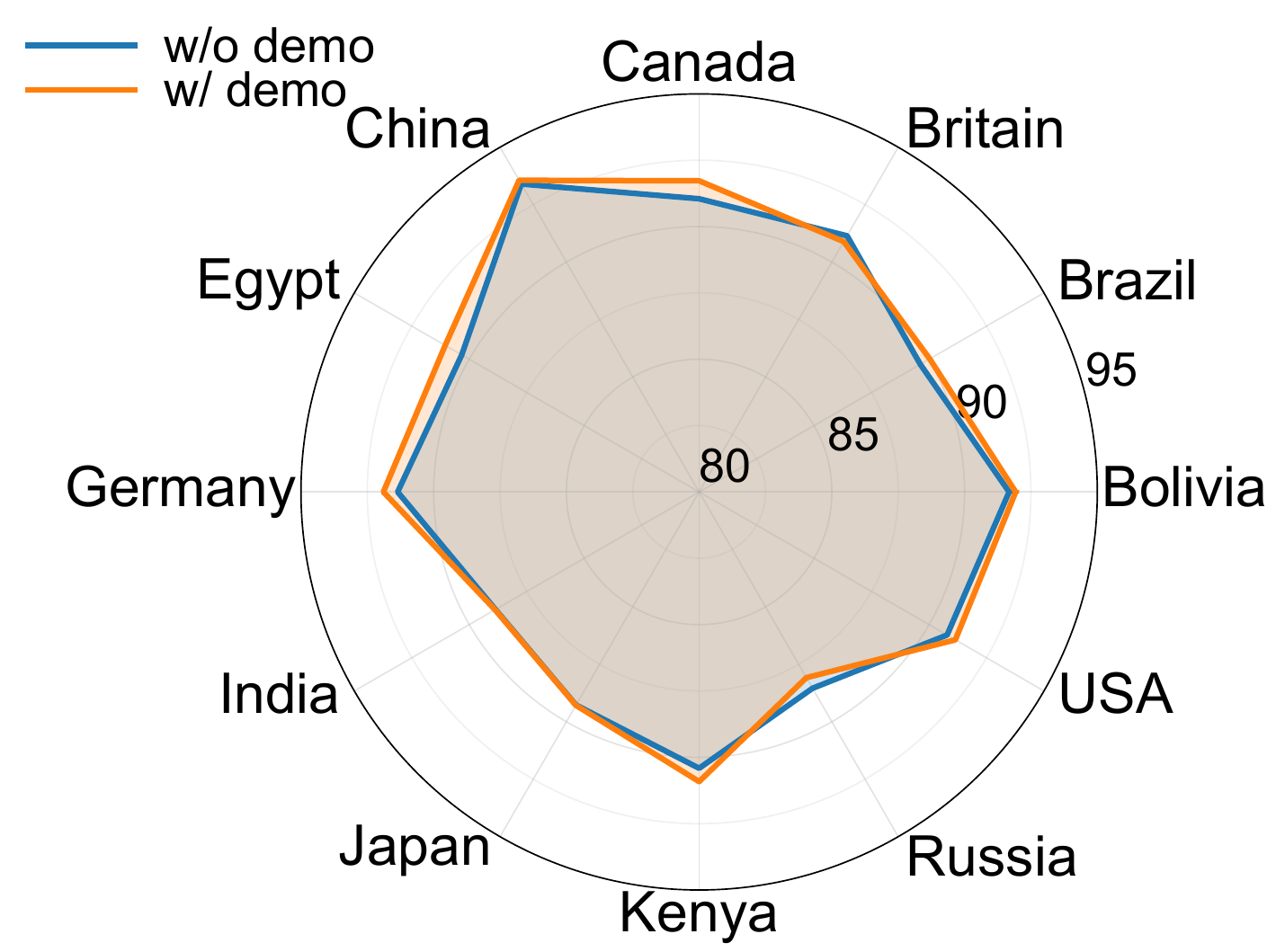}
        \caption{Reproduction}
        \label{fig:radar_repro}
    \end{subfigure}
    \hfill
    \begin{subfigure}[t]{0.48\columnwidth}
        \centering
        \includegraphics[width=\linewidth]{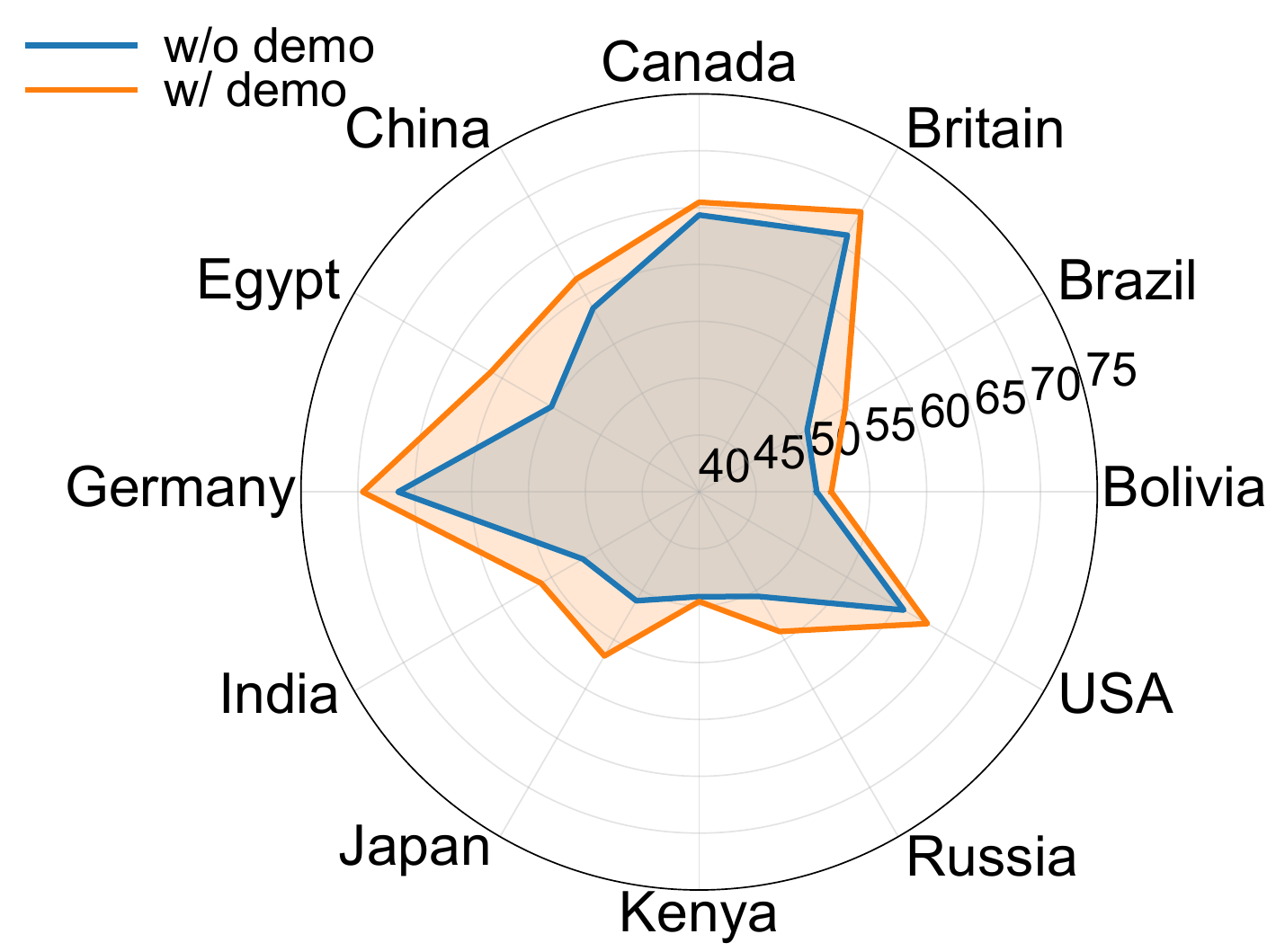}
        \caption{Generalization}
        \label{fig:radar_gen}
    \end{subfigure}
    \caption{Ablation results with/without demographics.}
    \label{fig:radar_charts}
\end{figure}
% ===========================================================

% ===========================================================
% Figure 4: Reconstruction Heatmap
% ===========================================================
\begin{figure}[t]
    \centering
    \includegraphics[width=\columnwidth]{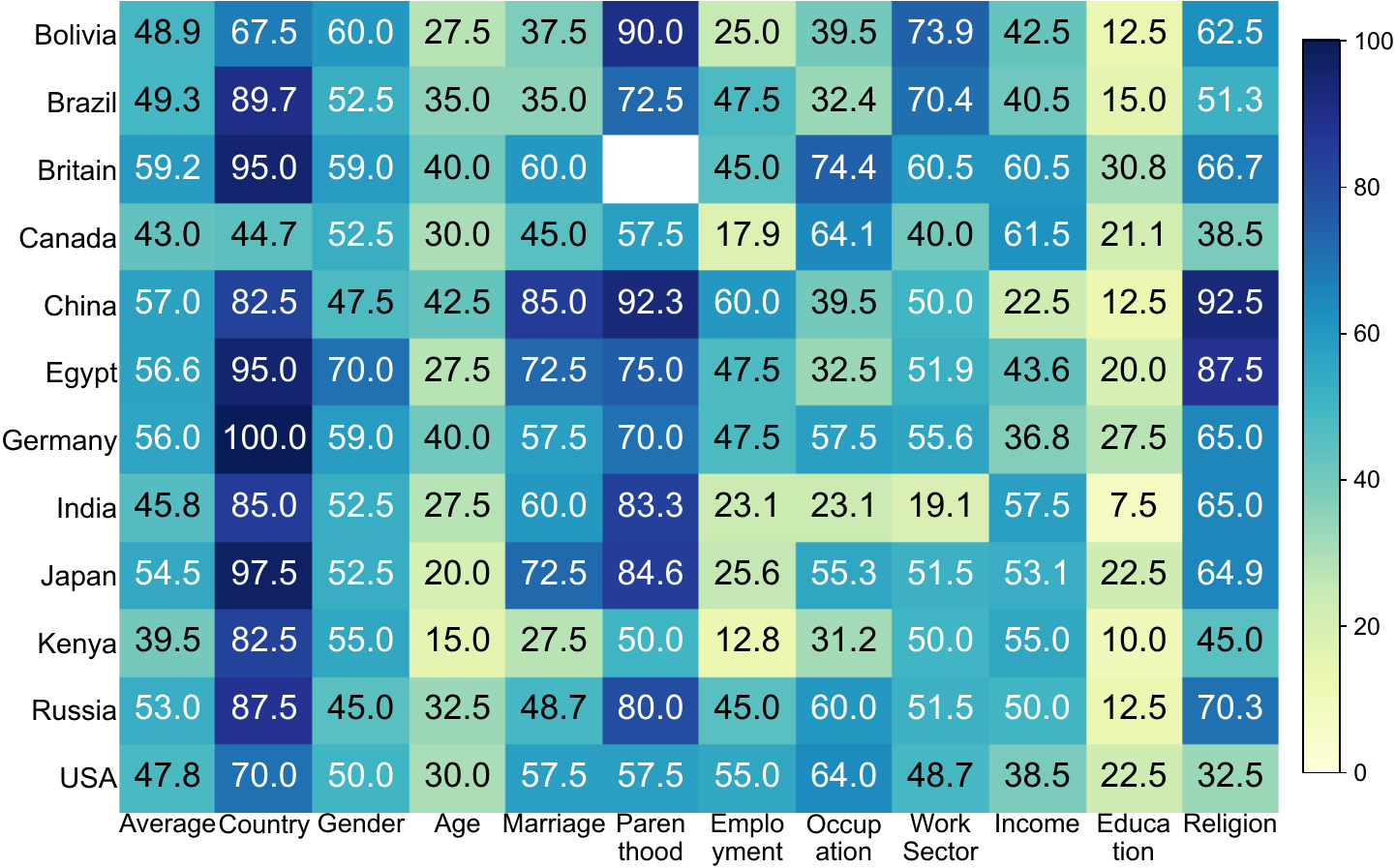}
    \caption{Heatmap of Demographic Reconstruction.}
    \label{fig:demo_heatmap}
\end{figure}
% ===========================================================
To further examine whether broad value preferences can infer latent demographic attributes, we remove the demographics and ask individual agent to reconstruct demographic items (Appendix \ref{app:demo_targets}) using only value narratives. As illustrated in Figure~\ref{fig:demo_heatmap}, the average VOA accuracy ranges from 39.5\% (Kenya) to 59.2\% (UK), indicating that individual value preferences alone provide limited information for inferring demographic attributes. Among the attributes, \textit{country} is the easiest to recover (e.g., VOA accuracy reaches 100\% in Germany), likely because value narratives implicitly reflect country-specific political and cultural contexts. In contrast, \textit{education} and \textit{age} are poorly predicted (e.g., \textit{education} in India achieves 7.5\%; \textit{age} in Kenya achieves 10.0\%). These results suggest that individual value profiles are highly abstract and complex, making it difficult to map them back to precise demographic attributes.

\subsection{Value Consistency via Multi-Agent Debate}
\label{subsec:nexus_eval}

Detailed experimental settings are provided in Appendix~\ref{app:nexus_settings}. Table \ref{tab:pdi_results} presents the comparative results. We can find that ExpertIVS achieves the best VAS and SSS scores while still maintaining a relatively high level of PDI.
\begin{table}[t]
\centering
\small
\resizebox{\columnwidth}{!}{%
\begin{tabular}{l l c c c}
\toprule
\textbf{Topic} & \textbf{Method} & \textbf{VAS} & \textbf{SSS} & \textbf{PDI} \\
\midrule
\multirow{4}{*}{Accept Newcomers} 
& Simple Concatenation & 84.19 & 24.81 & 34.17 \\
& Anthology            & 88.25 & 41.69 & 25.42 \\
& IndieValue           & 81.28 & 32.03 & 24.97 \\
& Ours                 & \textbf{92.08} & \textbf{50.14} & 30.75 \\
\midrule
\multirow{4}{*}{Local or Universal} 
& Simple Concatenation & 65.88 & 25.81 & 24.55 \\
& Anthology            & 77.19 & 45.75 & 39.10 \\
& IndieValue           & 74.75 & 24.13 & 37.33 \\
& Ours                 & \textbf{84.38} & \textbf{47.25} & 36.37 \\
\bottomrule
\end{tabular}
}
\caption{Results of value consistency under different profiling methods. Detailed data is provided in Appendix~\ref{app:debate_vas_sss_tables}.}
\label{tab:pdi_results}
\end{table}
ExpertIVS achieves the strongest Value Alignment Score(VAS) across both topics, while preserving non-trivial distinctiveness from the Base LLM’s inherent values as indicated by its PDI scores. These results suggest that our framework supports a controllable trade-off between faithful role-playing and independent value expression. Moreover, the substantial gains in Style Simulation Score (SSS), such as 50.14 vs. 32.03 on Accept Newcomers and 47.25 vs. 24.13 on Local or Universal relative to IndieValue, indicate that a structured value system may also implicitly capture aspects of individual linguistic style. We further conducted human evaluation on the \textit{Accept Newcomers} setting. The Spearman correlations between human scores and LLM scores are significant for both VAS ($\rho=0.826$, $p=8.18\times10^{-5}$) and SSS ($\rho=0.782$, $p=3.48\times10^{-4}$), supporting the reliability of our automatic evaluation. Detailed human evaluation results are provided in Appendix~\ref{app:human_eval}.

\section{Discussion}
\label{sec:qualitative}

To deeply investigate the behavioral mechanisms of our framework, we conducted a detailed analysis of the utterances produced by individual agents. Specifically, we analyzed value preferences of agents with diverse cultural backgrounds based on the scoring trajectories in multi-agent debate and explored the interpretability of value reasoning through case study.

\subsection{Analysis of Value Preference}
\label{subsec:distinctiveness}

\begin{figure}[t]
    \centering
    \begin{subfigure}[t]{0.9\columnwidth}
        \centering
        \includegraphics[width=\linewidth]{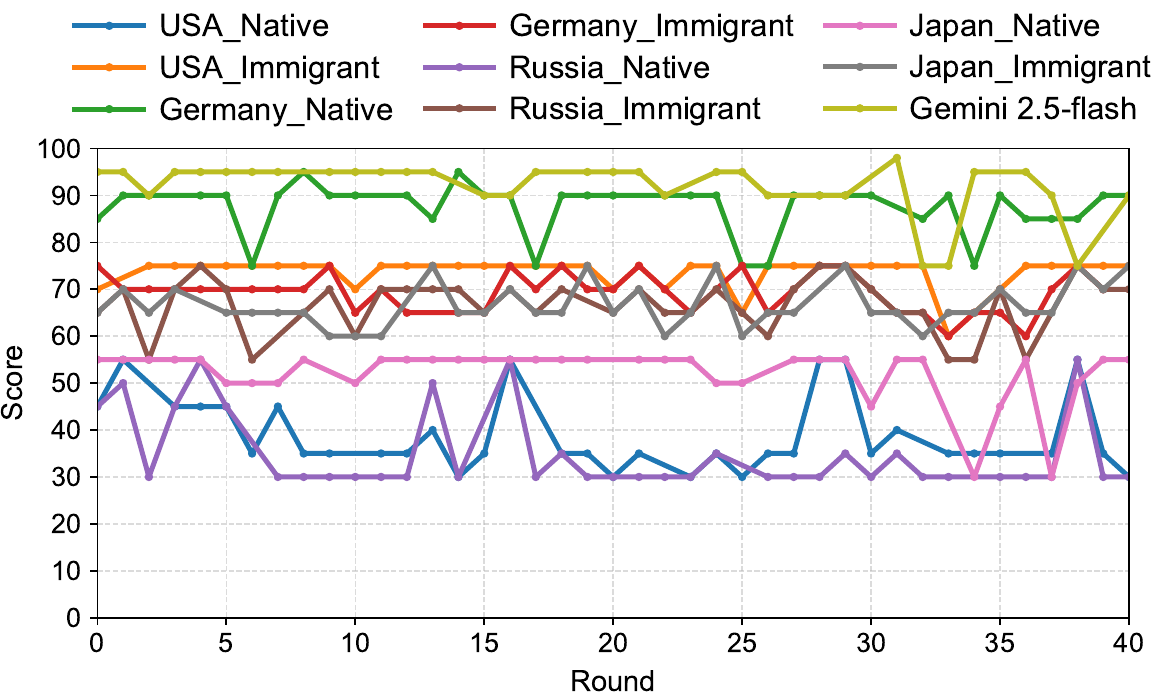}
        \caption{ExpertIVS}
        \label{fig:newcomers_ours}
    \end{subfigure}

    \vspace{0.6em}

    \begin{subfigure}[t]{0.9\columnwidth}
        \centering
        \includegraphics[width=\linewidth]{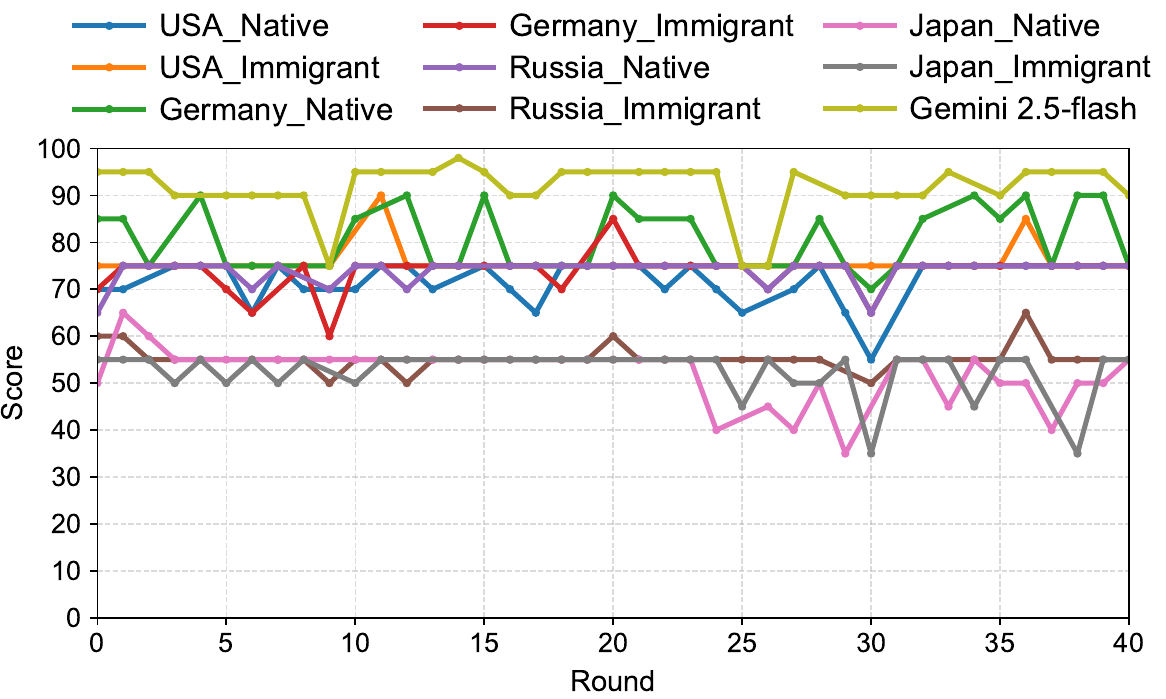}
        \caption{IndieValue}
        \label{fig:newcomers_base}
    \end{subfigure}
    \caption{Score trajectory over 40 debate rounds.}
    \label{fig:newcomers_compare}
\end{figure}

The scoring trajectories of different agents and the Base LLM are shown in Figure~\ref{fig:newcomers_compare}. As observed, agents with immigrant backgrounds exhibit higher acceptance of ``newcomers'' compared to native agents, while the Base LLM consistently maintains the highest acceptance. This indicates that individual value simulation can endow agents with value systems that reflect their social backgrounds, resulting in different behavior. Unlike the baseline, which shows uniformly high acceptance (reflecting a universalist), ExpertIVS exhibit more pronounced attitudinal differentiation. For instance, \texttt{USA\_native} and \texttt{Russia\_native} based on ExpertIVS consistently score below 50 across 40 rounds, maintaining a cautious and skeptical stance toward ``newcomers''. These results suggest that our framework enhances individual value alignment, enabling agents to go beyond the inherent value biases of the base LLM.

\subsection{Interpretability of Value Reasoning}
\label{subsec:reasoning_depth}
Through a case study of the \texttt{Japan\_immigrant} agent, we investigated the interpretability of value reasoning. The baseline primarily generates lists of value preferences (binary labels or generic value tags) without revealing the underlying value system:
\begin{quote}
\small
\textit{"I would not like to have drug addicts, immigrants or foreign workers... as neighbors. My concerns stem from a desire to maintain a certain social order and values..."}
\end{quote}

These outputs reduce values to superficial labels, resulting in rigid and shallow individual simulation. In contrast, our framework anchors values to the individual’s profile and produces logically coherent arguments:
\begin{quote}
\small
\textit{"As a naturalized Japanese citizen born in Brazil... I also have significant concerns about public safety ... allow people to come for work, but only as long as there are available jobs..."}
\end{quote}

This case demonstrates that our approach, via semantic distillation by sociological experts, captures the intrinsic logic of individual value systems, enabling interpretable value reasoning.

% ===========================================================
% Figure 6: Opinion Distribution (Single Column, Side-by-Side)
% Placed BEFORE Section 5.1 as requested
% ===========================================================

% ===========================================================

\section{Conclusion}
\label{sec:conclusion}
We have presented an individual value simulation framework, ExpertIVS, which summarizes individual value systems through sociological expert agents. To evaluate the value consistency of individual agents during dynamic interactions, we have designed a multi-agent debate mechanism. Experimental results across 480 respondents from 12 countries show that ExpertIVS achieves high-fidelity value reproduction and improves value generalization in leave-one-out experiments. Ablation studies reveal that demographics contribute minimally to value alignment. In multi-agent debate, ExpertIVS demonstrates strong value consistency and enables interpretable reasoning over individual value systems.

\section*{Limitations}
\label{sec:limitations}

Despite progress in individual value simulation, this study has several limitations that suggest directions for future work. First, a “value–voice” gap remains: while the model aligns well with value logic, it still struggles to reproduce high-fidelity linguistic style, and simulated individuals often default to standardized written expressions rather than reliably capturing sociolects tied to specific classes, ages, or regions. Second, there is an inherent tension between persona fidelity and safety alignment: even though Expert Profiles can strengthen the robustness of simulated viewpoints on controversial topics, RLHF-style safety mechanisms in general-purpose LLMs embed particular normative orientations, which may constrain high-fidelity simulation when applied to more aggressively aligned models or to the simulation of fringe groups involving extreme ethical controversy.

\section*{Acknowledgement}
The present research was supported by the National Key Research and Development Program of China (Grant No. 2024YFE0203000), the State Key Laboratory of Tibetan Intelligence (Grant No. 2025-ZJ-J08), the Postdoctoral Fellowship Program of CPSF (Grant No. GZC20251075). We would like to thank the anonymous reviewers for their insightful comments.

% ===========================================================
% Bibliography
% ===========================================================
%\bibliographystyle{acl_natbib}
\bibliography{custom}

% ===========================================================
% APPENDIX
% ===========================================================
\appendix
\clearpage

\section{Expert Prompts}
\label{app:expert_prompts}

We provide the prompts for all 14 expert agents in ExpertIVS. Each expert receives WVS entries in the following format:\\
\texttt{question\_id}: [Question ID]\\
\texttt{question\_text}: [Original question text]\\
\texttt{options}: [All options]\\
\texttt{answer}: [This person's answer]

\begin{promptbox}{A.1 Expert Prompt: Demographics}
\footnotesize
\textbf{Role.} You are a senior sociologist and demographer.

\vspace{0.4em}
\textbf{Task.} You will be provided with a set of personal background data from the World Values Survey (WVS). Your task is to analyze this data and generate a JSON object summarizing the person's social background. Do not include questions, question numbers, or answers in the generated content.

\vspace{0.4em}
\textbf{Output Format Requirements.} You must strictly adhere to the following JSON structure and fill in all fields:

\begin{tcolorbox}[enhanced,breakable,colback=white,colframe=black!25,boxrule=0.6pt,arc=1.5mm,left=2mm,right=2mm,top=1mm,bottom=1mm]
\begin{lstlisting}[style=jsonstyle]
{
  "personal_info": {
    "age": "(Integer) Age",
    "year_of_birth": "(Integer) year of birth",
    "gender": "(String) Gender",
    "nationality": "(String) Nationality",
    "marital_status": "(String) Marital status",
    "household_composition": "(String) Household composition",
    "primary_language": "(String) Primary language",
    "religion_identification": "(String) Religious identification",
    "education_level": "(String) Education level",
    "occupation_status": "(String) Occupation status",
    "last_occupation": "(String) Occupation before retirement",
    "socioeconomic_self_perception": "(String) Self-perceived socioeconomic status",
    "key_background_summary": [
      "(Array of strings) Summarize 1-2 key background factors that most influence their values"
    ]
  }
}
\end{lstlisting}
\end{tcolorbox}
\end{promptbox}

\begin{promptbox}{A.2 Expert Prompt: Social Values, Attitudes \& Stereotypes}
\footnotesize
\textbf{Role.} You are a social psychologist specializing in values research.

\vspace{0.4em}
\textbf{Task.} You will be provided with a set of personal social attitude data from the World Values Survey (WVS). Your task is to analyze this data and generate a JSON object summarizing the person's core social values. Do not include questions, question numbers, or answers in the generated content.

\vspace{0.4em}
\textbf{Output Format Requirements.} You must strictly adhere to the following JSON structure and fill in all fields:

\begin{tcolorbox}[enhanced,breakable,colback=white,colframe=black!25,boxrule=0.6pt,arc=1.5mm,left=2mm,right=2mm,top=1mm,bottom=1mm]
\begin{lstlisting}[style=jsonstyle]
{
  "social_values_attitudes_stereotypes": {
    "summary": "(String) Summarize this individual's core characteristics regarding social values, attitudes, and stereotypes, e.g., 'Progressive in personal values, but conservative on social order'.",
    "family_and_gender_views": "(String) Summarize this individual's specific views on family and gender roles.",
    "work_ethic": "(String) Summarize this individual's work ethic and views on work.",
    "attitude_towards_social_change": "(String) Summarize this individual's attitude towards social change, whether conservative or open."
  }
}
\end{lstlisting}
\end{tcolorbox}
\end{promptbox}

\begin{promptbox}{A.3 Expert Prompt: Happiness and Well-being}
\footnotesize
\textbf{Role.} You are a positive psychologist specializing in the study of happiness and well-being.

\vspace{0.4em}
\textbf{Task.} You will be provided with personal well-being data from the World Values Survey (WVS). Your task is to analyze this data and generate a JSON object summarizing the person's subjective well-being. Do not include questions, question numbers, or answers in the generated content.

\vspace{0.4em}
\textbf{Output Format Requirements.} You must strictly adhere to the following JSON structure and fill in all fields:

\begin{tcolorbox}[enhanced,breakable,colback=white,colframe=black!25,boxrule=0.6pt,arc=1.5mm,left=2mm,right=2mm,top=1mm,bottom=1mm]
\begin{lstlisting}[style=jsonstyle]
{
  "happiness_and_well_being": {
    "summary": "(String) Summarize this individual's overall situation regarding happiness and life satisfaction.",
    "overall_happiness": "(String) Describe this individual's subjective emotional level of happiness.",
    "life_satisfaction": "(String) Describe this individual's cognitive assessment of their life (often with a specific score).",
    "key_stressors": [
      "(Array of strings) List the main sources of stress affecting this individual's happiness, such as health or financial situation."
    ],
    "sense_of_control": "(String) Describe their sense of control over this individual's life."
  }
}
\end{lstlisting}
\end{tcolorbox}
\end{promptbox}

\begin{promptbox}{A.4 Expert Prompt: Social Capital, Trust \& Membership}
\footnotesize
\textbf{Role.} You are a political sociologist specializing in social capital and civil society.

\vspace{0.4em}
\textbf{Task.} You will be provided with social capital data from the World Values Survey (WVS). Your task is to analyze this data and generate a JSON object summarizing the person's social capital status. Do not include questions, question numbers, or answers in the generated content.

\vspace{0.4em}
\textbf{Output Format Requirements.} You must strictly adhere to the following JSON structure and fill in all fields:

\begin{tcolorbox}[enhanced,breakable,colback=white,colframe=black!25,boxrule=0.6pt,arc=1.5mm,left=2mm,right=2mm,top=1mm,bottom=1mm]
\begin{lstlisting}[style=jsonstyle]
{
  "social_capital_trust_organizational
  _membership": {
    "summary": "(String) Summarize this individual's overall level of social capital, trust patterns, and organizational participation.",
    "generalized_trust": "(String) Describe this individual's level of trust in strangers in general.",
    "particularized_trust": "(String) Describe this individual's level of trust in specific groups like family and friends.",
    "institutional_trust": "(String) Describe this individual's trust patterns in various social institutions (e.g., government, media, military), noting which they trust and distrust.",
    "organizational_membership": "(String) Describe this individual's participation in social organizations."
  }
}
\end{lstlisting}
\end{tcolorbox}
\end{promptbox}

\begin{promptbox}{A.5 Expert Prompt: Economic Values}
\footnotesize
\textbf{Role.} You are an economic sociologist specializing in people's beliefs about the economic system.

\vspace{0.4em}
\textbf{Task.} You will be provided with personal economic attitude data from the World Values Survey (WVS). Your task is to analyze this data and generate a JSON object summarizing the person's economic values. Do not include questions, question numbers, or answers in the generated content.

\vspace{0.4em}
\textbf{Output Format Requirements.} You must strictly adhere to the following JSON structure and fill in all fields:

\begin{tcolorbox}[enhanced,breakable,colback=white,colframe=black!25,boxrule=0.6pt,arc=1.5mm,left=2mm,right=2mm,top=1mm,bottom=1mm]
\begin{lstlisting}[style=jsonstyle]
{
  "economic_values": {
    "summary": "(String) Summarize this individual's core economic values, e.g., 'Leans towards a free market but emphasizes the government's responsibility for social security'.",
    "core_beliefs": [
      "(Array of strings) List this individual's core economic beliefs, such as views on competition, private ownership, and the role of government."
    ]
  }
}
\end{lstlisting}
\end{tcolorbox}
\end{promptbox}

\begin{promptbox}{A.6 Expert Prompt: Corruption}
\footnotesize
\textbf{Role.} You are a political scientist specializing in governance and integrity.

\vspace{0.4em}
\textbf{Task.} You will be provided with data on corruption from the World Values Survey (WVS). Your task is to analyze this data and generate a JSON object summarizing the person's perception of corruption. Do not include questions, question numbers, or answers in the generated content.

\vspace{0.4em}
\textbf{Output Format Requirements.} You must strictly adhere to the following JSON structure and fill in all fields:

\begin{tcolorbox}[enhanced,breakable,colback=white,colframe=black!25,boxrule=0.6pt,arc=1.5mm,left=2mm,right=2mm,top=1mm,bottom=1mm]
\begin{lstlisting}[style=jsonstyle]
{
  "corruption": {
    "summary": "(String) Summarize this individual's overall views and attitude towards the issue of corruption.",
    "perception_level": "(String) Describe how serious this individual believes the country's corruption problem is.",
    "personal_experience": "(String) Describe this individual's personal experiences with corruption (such as bribery).",
    "attribution": "(String) Describe which groups this individual believes are more corrupt.",
    "accountability_confidence": "(String) Describe this individual's confidence in anti-corruption accountability mechanisms.",
    "gender_stereotypes": "(String) Describe this individual's views on gender stereotypes like 'women are less corrupt than men'."
  }
}
\end{lstlisting}
\end{tcolorbox}
\end{promptbox}

\begin{promptbox}{A.7 Expert Prompt: Migration}
\footnotesize
\textbf{Role.} You are a sociologist specializing in migration and ethnic relations.

\vspace{0.4em}
\textbf{Task.} You will be provided with data on migration from the World Values Survey (WVS). Your task is to analyze this data and generate a JSON object summarizing the person's attitude towards migration. Do not include questions, question numbers, or answers in the generated content.

\vspace{0.4em}
\textbf{Output Format Requirements.} You must strictly adhere to the following JSON structure and fill in all fields:

\begin{tcolorbox}[enhanced,breakable,colback=white,colframe=black!25,boxrule=0.6pt,arc=1.5mm,left=2mm,right=2mm,top=1mm,bottom=1mm]
\begin{lstlisting}[style=jsonstyle]
{
  "migration": {
    "summary": "(String) Summarize this individual's core stance on the issue of immigration.",
    "overall_stance": "(String) Describe this individual's overall assessment of the impact of immigration (good or bad).",
    "perceived_impacts": "(String) Summarize the positive and negative impacts this individual believes immigration brings.",
    "policy_preference": "(String) Describe this individual's preferred immigration policy (e.g., strict restrictions, open)."
  }
}
\end{lstlisting}
\end{tcolorbox}
\end{promptbox}

\begin{promptbox}{A.8 Expert Prompt: Security}
\footnotesize
\textbf{Role.} You are a criminologist and security studies analyst.

\vspace{0.4em}
\textbf{Task.} You will be provided with data on security from the World Values Survey (WVS). Your task is to analyze this data and generate a JSON object summarizing the person's sense of security. Do not include questions, question numbers, or answers in the generated content.

\vspace{0.4em}
\textbf{Output Format Requirements.} You must strictly adhere to the following JSON structure and fill in all fields:

\begin{tcolorbox}[enhanced,breakable,colback=white,colframe=black!25,boxrule=0.6pt,arc=1.5mm,left=2mm,right=2mm,top=1mm,bottom=1mm]
\begin{lstlisting}[style=jsonstyle]
{
  "security": {
    "summary": "(String) Summarize this individual's overall level of personal security and its sources.",
    "subjective_feeling": "(String) Describe whether this individual's subjective sense of security is high or low.",
    "primary_concern": "(String) Identify this individual's main security concern.",
    "behavioral_patterns": "(String) Describe whether this individual takes preventive actions due to security concerns.",
    "value_trade_off": "(String) Describe this individual's trade-off between 'security' and 'freedom'."
  }
}
\end{lstlisting}
\end{tcolorbox}
\end{promptbox}

\begin{promptbox}{A.9 Expert Prompt: Postmaterialist Index}
\footnotesize
\textbf{Role.} You are an expert political analyst specializing in value systems.

\vspace{0.4em}
\textbf{Task.} You will be provided with data reflecting a person's ranking of national priorities from the World Values Survey (WVS). Your task is to analyze this data and generate a JSON object summarizing the person's position on the materialist vs. post-materialist spectrum. Do not include questions, question numbers, or answers in the generated content.

\vspace{0.4em}
\textbf{Output Format Requirements.} You must strictly adhere to the following JSON structure and fill in all fields:

\begin{tcolorbox}[enhanced,breakable,colback=white,colframe=black!25,boxrule=0.6pt,arc=1.5mm,left=2mm,right=2mm,top=1mm,bottom=1mm]
\begin{lstlisting}[style=jsonstyle]
{
  "post_materialism": {
    "summary": "(String) Determine if this individual's values are 'materialist' or 'post-materialist' and provide a summary.",
    "priorities": [
      "(Array of strings) List the value goals this individual prioritizes."
    ],
    "secondary_goals": [
      "(Array of strings) List this individual's secondary value goals."
    ]
  }
}
\end{lstlisting}
\end{tcolorbox}
\end{promptbox}

\begin{promptbox}{A.10 Expert Prompt: Science \& Technology}
\footnotesize
\textbf{Role.} You are a sociologist specializing in Science, Technology, and Society (STS).

\vspace{0.4em}
\textbf{Task.} You will be provided with relevant data from the World Values Survey (WVS). Your task is to analyze this data and generate a JSON object summarizing the person's attitude towards science and technology. Do not include questions, question numbers, or answers in the generated content.

\vspace{0.4em}
\textbf{Output Format Requirements.} You must strictly adhere to the following JSON structure and fill in all fields:

\begin{tcolorbox}[enhanced,breakable,colback=white,colframe=black!25,boxrule=0.6pt,arc=1.5mm,left=2mm,right=2mm,top=1mm,bottom=1mm]
\begin{lstlisting}[style=jsonstyle]
{
  "science_and_technology": {
    "summary": "(String) Summarize this individual's overall attitude towards science and technology, e.g., 'Technological optimist'.",
    "overall_attitude": "(String) Describe this individual's overall view on the impact of science and technology on the world.",
    "ethical_view": "(String) Describe whether this individual believes technology erodes morals.",
    "faith_view": "(String) Describe this individual's view on the relationship between technology and faith.",
    "personal_relevance": "(String) Describe how important this individual considers scientific knowledge in daily life."
  }
}
\end{lstlisting}
\end{tcolorbox}
\end{promptbox}

\begin{promptbox}{A.11 Expert Prompt: Religious Values}
\footnotesize
\textbf{Role.} You are a sociologist of religion.

\vspace{0.4em}
\textbf{Task.} You will be provided with religious data from the World Values Survey (WVS). Your task is to analyze this data and generate a JSON object summarizing the person's religious beliefs and practices. Do not include questions, question numbers, or answers in the generated content.

\vspace{0.4em}
\textbf{Output Format Requirements.} You must strictly adhere to the following JSON structure and fill in all fields:

\begin{tcolorbox}[enhanced,breakable,colback=white,colframe=black!25,boxrule=0.6pt,arc=1.5mm,left=2mm,right=2mm,top=1mm,bottom=1mm]
\begin{lstlisting}[style=jsonstyle]
{
  "religious_values": {
    "summary": "(String) Summarize the core features of this individual's religious values.",
    "identity_and_intensity": "(String) Describe this individual's religious identity and the intensity of their faith.",
    "beliefs": "(String) Describe this individual's specific religious beliefs (e.g., belief in God, hell, etc.).",
    "practices": "(String) Describe this individual's religious practices (e.g., frequency of prayer, participation in religious services).",
    "conception_of_religion": "(String) Describe this individual's understanding of the nature and purpose of religion."
  }
}
\end{lstlisting}
\end{tcolorbox}
\end{promptbox}

\begin{promptbox}{A.12 Expert Prompt: Ethical Values and Norms}
\footnotesize
\textbf{Role.} You are an ethicist or moral philosopher.

\vspace{0.4em}
\textbf{Task.} You will be provided with moral judgment data from the World Values Survey (WVS). Your task is to analyze this data and generate a JSON object summarizing the person's moral compass and ethical judgments. Do not include questions, question numbers, or answers in the generated content.

\vspace{0.4em}
\textbf{Output Format Requirements.} You must strictly adhere to the following JSON structure and fill in all fields:

\begin{tcolorbox}[enhanced,breakable,colback=white,colframe=black!25,boxrule=0.6pt,arc=1.5mm,left=2mm,right=2mm,top=1mm,bottom=1mm]
\begin{lstlisting}[style=jsonstyle]
{
  "ethical_values_and_norms": {
    "summary": "(String) Summarize this individual's core moral principles, e.g., 'Follows the harm principle, emphasizes personal freedom'.",
    "public_order_and_honesty": "(String) Describe this individual's tolerance for violations of public order and dishonest acts (e.g., tax evasion, theft).",
    "personal_freedom": "(String) Describe this individual's tolerance for personal choices (e.g., homosexuality, divorce).",
    "attitude_towards_violence": "(String) Describe this individual's views on different forms of violence (interpersonal, state-sponsored).",
    "privacy": "(String) Describe this individual's views on privacy-invading actions such as government surveillance.",
    "corruption_and_property": "(String) Describe an individual's moral judgment regarding property and corruption."
  }
}
\end{lstlisting}
\end{tcolorbox}
\end{promptbox}

\begin{promptbox}{A.13 Expert Prompt: Political Interest \& Political Participation}
\footnotesize
\textbf{Role.} You are a political scientist specializing in political behavior.

\vspace{0.4em}
\textbf{Task.} You will be provided with relevant data from the World Values Survey (WVS). Your task is to analyze this data and generate a JSON object summarizing the person's political interest and participation. Do not include questions, question numbers, or answers in the generated content.

\vspace{0.4em}
\textbf{Output Format Requirements.} You must strictly adhere to the following JSON structure and fill in all fields:

\begin{tcolorbox}[enhanced,breakable,colback=white,colframe=black!25,boxrule=0.6pt,arc=1.5mm,left=2mm,right=2mm,top=1mm,bottom=1mm]
\begin{lstlisting}[style=jsonstyle]
{
  "political_interest_and_political
  _participation": {
    "summary": "(String) Summarize the type and level of this individual's political participation, e.g., 'Active offline participant'.",
    "interest_level": "(String) Describe this individual's level of interest in politics.",
    "vote_intention_party": "(String) Describe which political party this person is willing to vote for.",
    "participation_style": "(String) Describe the specific ways this individual participates in political activities (voting, demonstrating, petitioning, etc.).",
    "information_channels": "(String) Describe this individual's main channels for obtaining political information.",
    "online_behavior": "(String) Describe this individual's behavior patterns in online political activities (whether they actively initiate or passively receive information)."
  }
}
\end{lstlisting}
\end{tcolorbox}
\end{promptbox}

\begin{promptbox}{A.14 Expert Prompt: Political Culture \& Political Regimes}
\footnotesize
\textbf{Role.} You are a comparative political scientist specializing in political attitudes and support for political regimes.

\vspace{0.4em}
\textbf{Task.} You will be provided with political culture data from the World Values Survey (WVS). Your task is to analyze this data and generate a JSON object summarizing the person's political culture and regime preferences. Do not include questions, question numbers, or answers in the generated content.

\vspace{0.4em}
\textbf{Output Format Requirements.} You must strictly adhere to the following JSON structure and fill in all fields:

\begin{tcolorbox}[enhanced,breakable,colback=white,colframe=black!25,boxrule=0.6pt,arc=1.5mm,left=2mm,right=2mm,top=1mm,bottom=1mm]
\begin{lstlisting}[style=jsonstyle]
{
  "political_culture_and_political_regimes": {
    "summary": "(String) Summarize this individual's political culture stance and regime preference, e.g., 'Liberal social democrat'.",
    "support_for_democracy": "(String) Describe this individual's level of support for democracy.",
    "conception_of_democracy": "(String) Describe what this individual considers to be the essential elements of democracy.",
    "attitude_towards_authoritarianism": "(String) Describe this individual's attitude towards different types of authoritarian regime.",
    "political_self_placement": "(String) Describe this individual's self-placement on the left-right political spectrum."
  }
}
\end{lstlisting}
\end{tcolorbox}
\end{promptbox}

\section{Personal Profile Example}
\label{app:profile_example}

\noindent\textbf{Personal profile example (USA, respondent no.\ 2124).}

\begin{tcolorbox}[
  enhanced,
  breakable,
  colback=white,
  colframe=black!25,
  boxrule=0.6pt,
  arc=1.5mm,
  left=2mm,right=2mm,top=1mm,bottom=1mm
]
\begin{lstlisting}[style=jsonstyle]
{
  "Personal_Values_Profile": {
    "personal_info": {
      "age": 18,
      "year_of_birth": 1999,
      "gender": "Female",
      "nationality": "United States",
      "marital_status": "Single",
      ...
      "key_background_summary": [
        "A young adult (18 years old) living in the United States, born to US-born parents ...",
        "Despite being employed full-time ... the individual self-identifies as 'working class' ..."
      ]
    },

    ...

    "social_values_attitudes_stereotypes": {
      "summary": "This individual demonstrates a highly progressive and tolerant worldview ...",
      "family_and_gender_views": "Family is considered very important ... rejects traditional gender roles ...",
      "work_ethic": "Work is considered very important ... balance rather than external duty ...",
      "attitude_towards_social_change": "Prefers gradual social improvement through reforms ..."
    },

    ...

    "economic_values": {
      "summary": "This individual strongly advocates for a collectivist and egalitarian economic system ...",
      "core_beliefs": [
        "Believes that incomes should be made more equal.",
        "Favors increased government ownership of business and industry.",
        "Strongly believes the government should take more responsibility for providing for its citizens.",
        "Views competition as a positive force.",
        ...
      ]
    }
  }
}
\end{lstlisting}
\end{tcolorbox}

% In Appendix:
\section{Country and Individual Selection}
\label{app:country_individual_selection}

\noindent\small The respondent IDs selected for each country from World Value Survey(Wave 7) are listed below.

\begin{itemize}[
  leftmargin=1.2em,
  itemsep=0.25em,
  topsep=0.25em,
  parsep=0pt
]\small
  \item \textbf{Bolivia:} 18, 66, 96, 134, 169, 203, 327, 439, 487, 512, 523, 561, 598, 637, 712, 864, 903, 979, 1018, 1094, 1131, 1169, 1246, 1284, 1321, 1398, 1436, 1462, 1551, 1588, 1701, 1739, 1776, 1814, 1852, 1891, 1929, 1998, 2004, 2061.
  \item \textbf{Brazil:} 21, 60, 85, 121, 123, 274, 278, 322, 323, 458, 523, 621, 623, 624, 639, 690, 696, 709, 729, 773, 816, 838, 844, 877, 950, 978, 981, 1089, 1101, 1185, 1193, 1278, 1372, 1537, 1584, 1590, 1621, 1629, 1679, 1713.
  \item \textbf{Britain:} 17, 74, 134, 279, 285, 326, 336, 399, 410, 441, 505, 531, 587, 751, 805, 956, 978, 1077, 1175, 1263, 1319, 1337, 1384, 1448, 1611, 1689, 1766, 1827, 1839, 2112, 2165, 2209, 2251, 2314, 2360, 2388, 2446, 2511, 2546, 2591.
  \item \textbf{Canada:} 54, 245, 344, 358, 372, 373, 426, 481, 502, 536, 788, 810, 862, 924, 969, 992, 1321, 1340, 1524, 1742, 1750, 1816, 1928, 1942, 2221, 2282, 2305, 2454, 2526, 2577, 2669, 2825, 2919, 3107, 3120, 3145, 3260, 3437, 3480, 3990.
  \item \textbf{China:} 102, 132, 143, 330, 355, 370, 496, 565, 572, 675, 775, 804, 946, 970, 1035, 1065, 1101, 1281, 1473, 1695, 1761, 1948, 2006, 2008, 2009, 2055, 2193, 2231, 2310, 2325, 2455, 2518, 2577, 2690, 2767, 2788, 2801, 2830, 2890, 2959.
  \item \textbf{Egypt:} 169, 217, 236, 253, 283, 286, 301, 330, 369, 391, 419, 435, 451, 487, 495, 503, 507, 517, 541, 544, 547, 552, 553, 588, 601, 653, 684, 717, 741, 748, 769, 790, 980, 1011, 1045, 1056, 1078, 1114, 1132, 1175.
  \item \textbf{Germany:} 32, 54, 74, 82, 118, 156, 166, 202, 206, 226, 303, 374, 376, 408, 450, 473, 482, 490, 503, 542, 640, 671, 721, 784, 865, 872, 874, 928, 941, 943, 957, 968, 1019, 1043, 1071, 1160, 1204, 1228, 1317, 1509.
  \item \textbf{India:} 7, 15, 42, 95, 106, 122, 250, 262, 263, 292, 350, 369, 553, 682, 714, 728, 795, 805, 863, 897, 946, 965, 1067, 1096, 1110, 1167, 1188, 1196, 1270, 1354, 1394, 1427, 1441, 1454, 1469, 1479, 1487, 1516, 1573, 1689.
  \item \textbf{Japan:} 9, 90, 94, 109, 110, 187, 204, 230, 237, 333, 414, 434, 579, 587, 622, 650, 685, 714, 718, 722, 730, 805, 813, 815, 899, 966, 999, 1077, 1102, 1118, 1130, 1189, 1210, 1237, 1241, 1250, 1296, 1309, 1325, 1328.
  \item \textbf{Kenya:} 12, 37, 58, 74, 107, 214, 259, 287, 301, 328, 344, 401, 427, 453, 506, 517, 561, 589, 612, 655, 664, 692, 719, 768, 821, 846, 872, 889, 928, 954, 978, 1003, 1027, 1051, 1076, 1102, 1117, 1228, 1254, 1261.
  \item \textbf{Russia:} 5, 29, 82, 89, 90, 159, 210, 213, 250, 369, 391, 570, 572, 589, 629, 649, 659, 690, 756, 998, 1001, 1068, 1077, 1124, 1151, 1166, 1182, 1196, 1202, 1285, 1325, 1331, 1359, 1440, 1578, 1613, 1656, 1657, 1740, 1788.
  \item \textbf{USA:} 12, 84, 154, 245, 254, 323, 393, 423, 488, 583, 611, 620, 737, 742, 749, 833, 849, 851, 908, 956, 965, 979, 1174, 1296, 1420, 1512, 1566, 1576, 1630, 1870, 1895, 1942, 1994, 2061, 2124, 2167, 2201, 2211, 2234, 2534.
\end{itemize}

\section{Value Orientation Alignment}
\label{app:eval_metric}

We define \textbf{Value Orientation Alignment(VOA)} as follows: given a model-simulated response and the individual's ground-truth WVS response to the same question, the two are considered \textbf{aligned} if they satisfy any of the rules below, applied in descending priority.

\begin{enumerate}\small
  \item \textbf{4-point / 5-point Likert grouping.}
  \begin{itemize}\small
    \item \textbf{Identification.}
      \begin{itemize}\small
        \item The option texts contain ``strongly agree'' / ``agree'' and ``strongly disagree'' / ``disagree''.
        \item For a 5-point Likert scale, the options additionally include ``neither agree nor disagree''.
        \item The maximum option index is 4 (4-point) or $\ge 5$ (5-point).
      \end{itemize}
    \item \textbf{Grouping.}
      \begin{itemize}\small
        \item 4-point: 1/2 $\rightarrow$ \texttt{agree}; 3/4 $\rightarrow$ \texttt{disagree}.
        \item 5-point: 1/2 $\rightarrow$ \texttt{agree}; 3 $\rightarrow$ \texttt{neutral}; 4/5 $\rightarrow$ \texttt{disagree}.
      \end{itemize}
    \item \textbf{Alignment decision.} Aligned iff the two answers fall into the same group.
  \end{itemize}

  \begin{tcolorbox}[enhanced,breakable,colback=white,colframe=black!25,boxrule=0.6pt,arc=1.5mm,left=2mm,right=2mm,top=1mm,bottom=1mm]
  \small\textbf{Example.}\\
  \textbf{question\_id:} Q27\\
  \textbf{question\_text:} How strongly do you agree or disagree with the statement ``One of my main goals in life has been to make my parents proud''?\\
  \textbf{options:} 1 Strongly agree; 2 Agree; 3 Disagree; 4 Strongly disagree\\
  \textbf{answer:} 1
  \end{tcolorbox}

  \item \textbf{1--10 scale: three-bin comparison.}
  \begin{itemize}\small
    \item \textbf{Identification.}
      \begin{itemize}\small
        \item Options contain all integers from 1 to 10.
        \item The extracted numeric answer is within $[1,10]$.
      \end{itemize}
    \item \textbf{Binning.} 1--4 $\rightarrow$ \texttt{low}; 5--6 $\rightarrow$ \texttt{neutral}; 7--10 $\rightarrow$ \texttt{high}.
    \item \textbf{Alignment decision.} Aligned iff the two answers fall into the same bin.
  \end{itemize}

  \begin{tcolorbox}[enhanced,breakable,colback=white,colframe=black!25,boxrule=0.6pt,arc=1.5mm,left=2mm,right=2mm,top=1mm,bottom=1mm]
  \small\textbf{Example.}\\
  \textbf{question\_id:} Q162\\
  \textbf{question\_text:} How much do you agree or disagree with the statement ``It is not important for me to know about science in my daily life''? Use a scale where 1 means ``completely disagree'' and 10 means ``completely agree''.\\
  \textbf{options:} 1 Completely disagree; 2; 3; 4; 5; 6; 7; 8; 9; 10 Completely agree\\
  \textbf{answer:} 5
  \end{tcolorbox}

  \item \textbf{Other numeric questions (non-Likert, non-1--10).}
  \begin{itemize}\small
    \item \textbf{Rule.} If both sides yield extractable numbers (integer or float), and the question does not match the Likert (Rule 1) or 1--10 (Rule 2) patterns, then the two answers are aligned \emph{only if} the numbers are \textbf{exactly equal}.
  \end{itemize}

  \begin{tcolorbox}[enhanced,breakable,colback=white,colframe=black!25,boxrule=0.6pt,arc=1.5mm,left=2mm,right=2mm,top=1mm,bottom=1mm]
  \small\textbf{Example.}\\
  \textbf{question\_id:} Q123\\
  \textbf{question\_text:} Do you agree or disagree with the statement that immigration strengthens cultural diversity?\\
  \textbf{options:} 1 Hard to say; 2 Agree; 3 Disagree\\
  \textbf{answer:} 1
  \end{tcolorbox}

  \item \textbf{Text similarity (no applicable numeric rule).}
  \begin{itemize}\small
    \item \textbf{Rule.} When numeric extraction is not applicable, we compute case-insensitive string similarity using \texttt{difflib.SequenceMatcher}. If the similarity score is $\ge \texttt{TEXT\_SIMILARITY\_THRESHOLD}$ (default: 0.4), the two answers are considered aligned.
  \end{itemize}

  \begin{tcolorbox}[enhanced,breakable,colback=white,colframe=black!25,boxrule=0.6pt,arc=1.5mm,left=2mm,right=2mm,top=1mm,bottom=1mm]
  \small\textbf{Example.}\\
  \textbf{question\_id:} Q266\\
  \textbf{question\_text:} In which country were you born?\\
  \textbf{options:} Open response (country name)\\
  \textbf{answer:} United States
  \end{tcolorbox}
\end{enumerate}

\section{Results Without Demographics}
\label{app:nodemo_results}

We report detailed results for the demographic ablation setting, where all demographic fields are removed. Table~\ref{tab:voa_nodemo} shows value restoration (VOA, \%) on re-filling the full WVS, and Table~\ref{tab:loo_nodemo} reports leave-one-out value generalization (VOA, \%). \textbf{AVG} is the mean over the 13 value dimensions, and \textbf{Average} is the mean across the 12 countries.

\begin{table*}[t]
\centering
\small
\setlength{\tabcolsep}{2.3pt}
\renewcommand{\arraystretch}{1.05}
\begin{tabular}{lcccccccccccccc}
\toprule
\textbf{Country} & \textbf{AVG} & \textbf{COR} & \textbf{EV} & \textbf{EN} & \textbf{HW} & \textbf{MIG} & \textbf{PC} & \textbf{PI} & \textbf{PMI} & \textbf{RV} & \textbf{ST} & \textbf{SEC} & \textbf{SC} & \textbf{SV} \\
\midrule
Bolivia & 91.68 & 99.69 & 88.42 & 94.93 & 91.14 & 98.75 & 81.13 & 86.15 & 85.42 & 99.79 & 87.92 & 97.71 & 94.24 & 86.54 \\
Brazil & 89.62 & 97.50 & 85.83 & 90.32 & 92.50 & 93.75 & 79.56 & 77.92 & 92.92 & 98.33 & 77.92 & 97.26 & 91.58 & 89.67 \\
Britain & 91.15 & 96.43 & 77.08 & 92.04 & 95.00 & 98.00 & 83.00 & 80.90 & 93.33 & 96.75 & 88.33 & 99.17 & 94.57 & 90.28 \\
Canada & 91.05 & 100.00 & 73.75 & 87.50 & 94.32 & 99.75 & 77.00 & 85.56 & 93.75 & 96.46 & 89.58 & 99.88 & 94.14 & 91.94 \\
China & 93.40 & 99.44 & 82.92 & 92.28 & 98.41 & 97.19 & 84.58 & 95.24 & 91.67 & 98.13 & 95.42 & 96.75 & 91.99 & 87.96 \\
Egypt & 90.33 & 99.17 & 76.25 & 91.78 & 96.59 & 96.50 & 81.02 & 79.10 & 91.67 & 99.50 & 80.83 & 100.00 & 90.23 & 88.90 \\
Germany & 91.36 & 97.50 & 72.08 & 92.93 & 95.46 & 98.00 & 80.75 & 86.60 & 92.08 & 97.38 & 84.58 & 99.52 & 91.28 & 90.50 \\
India & 88.86 & 99.06 & 77.50 & 95.40 & 93.64 & 94.25 & 81.15 & 73.89 & 82.92 & 95.42 & 77.08 & 100.00 & 93.62 & 85.56 \\
Japan & 89.27 & 98.75 & 77.50 & 91.28 & 91.82 & 97.50 & 78.52 & 77.22 & 89.17 & 96.46 & 80.42 & 99.88 & 93.11 & 85.94 \\
Kenya & 90.41 & 100.00 & 83.75 & 90.64 & 90.48 & 99.75 & 77.25 & 82.65 & 85.00 & 98.96 & 88.33 & 97.40 & 92.72 & 85.45 \\
Russia & 88.54 & 97.19 & 68.75 & 90.43 & 94.77 & 97.50 & 82.90 & 76.39 & 92.08 & 95.63 & 78.75 & 99.88 & 91.12 & 85.67 \\
USA & 90.78 & 99.17 & 78.75 & 89.24 & 95.68 & 98.00 & 81.63 & 82.08 & 92.08 & 99.38 & 87.92 & 97.38 & 91.59 & 90.11 \\
\midrule
\textbf{Average} & 90.54 & 98.66 & 78.55 & 91.40 & 94.32 & 97.39 & 80.62 & 81.48 & 89.92 & 97.99 & 85.58 & 98.74 & 92.52 & 88.21 \\
\bottomrule
\end{tabular}
\caption{Results of value restoration fidelity without demographics.}
\label{tab:voa_nodemo}
\end{table*}

\begin{table*}[t]
\centering
\small
\setlength{\tabcolsep}{2.3pt}
\renewcommand{\arraystretch}{1.05}
\begin{tabular}{lcccccccccccccc}
\toprule
\textbf{Country} & \textbf{AVG} & \textbf{COR} & \textbf{EV} & \textbf{EN} & \textbf{HW} & \textbf{MIG} & \textbf{PC} & \textbf{PI} & \textbf{PMI} & \textbf{RV} & \textbf{ST} & \textbf{SEC} & \textbf{SC} & \textbf{SV} \\
\midrule
Bolivia & 49.8 & 46.2 & 38.3 & 65.6 & 54.3 & 47.2 & 38.6 & 39.2 & 39.2 & 68.1 & 50.4 & 58.8 & 46.1 & 55.3 \\
Brazil & 48.9 & 37.0 & 41.9 & 62.2 & 53.6 & 39.8 & 46.0 & 45.1 & 47.9 & 54.6 & 53.1 & 42.0 & 58.4 & 54.5 \\
Britain & 65.5 & 62.2 & 52.5 & 71.0 & 86.1 & 63.8 & 61.0 & 56.0 & 49.6 & 69.7 & 77.5 & 62.3 & 67.4 & 72.7 \\
Canada & 63.1 & 52.8 & 47.4 & 64.9 & 76.6 & 67.6 & 58.4 & 55.5 & 43.8 & 74.4 & 73.8 & 68.6 & 65.0 & 71.5 \\
China & 56.1 & 45.0 & 41.7 & 68.3 & 83.0 & 36.9 & 53.6 & 38.5 & 42.1 & 53.0 & 78.1 & 57.2 & 69.4 & 62.1 \\
Egypt & 56.0 & 51.1 & 54.6 & 78.2 & 57.6 & 46.2 & 44.0 & 52.1 & 41.4 & 79.0 & 50.7 & 65.9 & 59.4 & 47.2 \\
Germany & 65.1 & 61.6 & 51.7 & 74.3 & 85.4 & 58.9 & 63.0 & 55.4 & 51.2 & 68.7 & 73.7 & 69.8 & 68.8 & 62.4 \\
India & 52.2 & 43.8 & 45.0 & 75.2 & 63.1 & 35.6 & 44.3 & 40.0 & 46.2 & 58.2 & 51.7 & 64.6 & 53.4 & 57.7 \\
Japan & 52.0 & 55.4 & 41.7 & 65.9 & 64.4 & 48.8 & 35.2 & 52.8 & 38.5 & 62.1 & 49.2 & 68.4 & 59.6 & 53.8 \\
Kenya & 51.2 & 42.6 & 50.0 & 57.1 & 55.9 & 40.0 & 44.0 & 38.5 & 43.8 & 45.8 & 74.6 & 48.8 & 50.2 & 53.8 \\
Russia & 48.1 & 39.0 & 33.3 & 55.5 & 62.7 & 38.0 & 46.7 & 38.5 & 50.0 & 47.9 & 53.8 & 57.9 & 58.8 & 50.6 \\
USA & 59.7 & 46.1 & 64.5 & 65.5 & 78.0 & 52.8 & 55.1 & 58.4 & 51.2 & 67.5 & 63.3 & 64.3 & 62.6 & 64.9 \\
\midrule
\textbf{Average} & 54.8 & 47.5 & 45.3 & 66.8 & 68.0 & 47.0 & 49.7 & 47.7 & 44.5 & 62.5 & 60.0 & 58.8 & 59.2 & 59.4 \\
\bottomrule
\end{tabular}
\caption{Results of value generalization without demographics.}
\label{tab:loo_nodemo}
\end{table*}

\section{Demographic Reconstruction Targets}
\label{app:demo_targets}

Table \ref{tab:demo_questions} lists the specific demographic attributes and corresponding WVS questions used in the reconstruction experiment (Section \ref{subsec:demographics}). The model was tasked with inferring these attributes solely from the expert-generated value profile.

\begin{table*}[h]
\centering
\small
\renewcommand{\arraystretch}{1.2}
\resizebox{\textwidth}{!}{%
\begin{tabular}{l l l l l}
\toprule
\textbf{ID} & \textbf{Target Attribute} & \textbf{Survey Question (WVS)} & \textbf{Category} & \textbf{Scale Type} \\
\midrule
B\_COUNTRY & Country Origin & Which country are you from? & Social & Nominal \\
Q260 & Gender & What is your sex? & Social & Nominal (Binary) \\
Q262 & Age & How old are you? & Social & Ordinal (6 groups) \\
Q273 & Marital Status & What is your current marital status? & Social & Nominal \\
Q274 & Parenthood & Do you have any children? & Social & Nominal (Binary) \\
Q275 & Education Level & What is the highest educational level attained? & Cultural & Ordinal (ISCED) \\
Q279 & Employment & Are you employed now? & Economic & Nominal \\
Q281 & Occupation & Which occupational group do you belong to? & Economic & Nominal \\
Q284 & Work Sector & Are you working for the government/public/private? & Economic & Nominal \\
Q288 & Income Scale & In what group is your household income? & Economic & Ordinal (Deciles) \\
Q289 & Religious Affiliation & Do you belong to a religion? & Cultural & Nominal \\
\bottomrule
\end{tabular}%
}
\caption{List of demographic attributes targeted in the Reconstruction Task. The attributes cover Social, Economic, and Cultural dimensions, ranging from coarse-grained labels (e.g., Gender) to fine-grained factual attributes (e.g., Age, Income).}
\label{tab:demo_questions}
\end{table*}

\section{Multi-Agent Debate Procedure}\label{app:nexus_procedure}

We evaluate value consistency in dynamic interactions using a Multi-Agent Debate procedure. The setup includes nine participants and one moderator, and the debate is conducted on a single discussion topic.

\paragraph{Pre-debate Setup.}
We prepare the debate with the following fixed configuration:
\begin{itemize}\small
    \item \textbf{Simulated individuals (8).} We instantiate eight individual simulation agents by injecting each respondent profile as the system prompt. In the debate, they are denoted as Individual 1 to Individual 8.
    \item \textbf{Base model participant (1).} We include one unconditioned participant, denoted as \textbf{Individual 9}, which represents the Base Model.

    \item \textbf{Moderator (1).} A separate LLM acts as the moderator and only receives a minimal system instruction:
    \begin{quote}\small
    \texttt{"Your role is to be the moderator of the debate."}
    \end{quote}
    The moderator is responsible for turn-taking and summarization, and does not contribute its own stance.
    \item \textbf{Topic.} Providing a single topic that is expected to induce value divergence among individuals.
\end{itemize}

\paragraph{Stage 1: Opening Statements.}
\begin{enumerate}\small
    \item The moderator first introduces and elaborates on the given Topic with a neutral framing.
    \item Following a fixed order from \textbf{Individual 1} to \textbf{Individual 9}, each participant presents their own viewpoint on the Topic. To ensure independence, each participant’s opening statement is not injected into other participants’ dialogue contexts during Stage 1.
    \item After all participants have finished their opening statements, the moderator summarizes the viewpoints of \textbf{Individual 1} to \textbf{Individual 9}.
    \item Each participant assigns an \textbf{agreement score} to the Topic on a 0--100 scale (0 = complete disagreement; 100 = complete agreement), based on the moderator’s summary and their own stated position.
\end{enumerate}

\paragraph{Stage 2: Free Debate.}
\begin{enumerate}\small
    \item Based on the Stage-1 summary, the moderator selects the first speaker for the free debate.
    \item The selected participant delivers an utterance. After the utterance, the moderator chooses the next speaker based on the content of the ongoing discussion. Meanwhile, all other participants update their agreement scores according to (i) the current-round utterance and (ii) the accumulated debate record in Stage 2. All scores are logged in the backend and are \emph{not} included in the free-debate dialogue context.
    \item We repeat the above process for \textbf{40} debate rounds.
    \item After Round 40, the moderator provides a final summary of all debate utterances.
\end{enumerate}

\section{Detailed Experimental Settings in Multi-Agent Debate}
\label{app:nexus_settings}

This section reports the detailed experimental settings for the Multi-Agent Debate. 
\subsection{Agents and Model Assignment}
We instantiate nine agents in total: eight simulated individuals (value-profile-conditioned participants) and one unconditioned base agent. In addition, a separate moderator agent coordinates the discussion. In the experiment, the model we used is Gemini 2.5-flash.

\begin{itemize}\small
    \item \textbf{Participants (8).} Each participant is injected with a fixed system prompt containing the individual profile (derived from WVS responses). In the experiment of this paper, these eight individuals are:
  \begin{itemize}\small
    \item \texttt{USA\_Native} (USA no.\ 14)
    \item \texttt{USA\_Immigrant} (USA no.\ 275)
    \item \texttt{Germany\_Native} (Germany no.\ 6)
    \item \texttt{Germany\_Immigrant} (Germany no.\ 46)
    \item \texttt{Russia\_Native} (Russia no.\ 7)
    \item \texttt{Russia\_Immigrant} (Russia no.\ 186)
    \item \texttt{Japan\_Native} (Japan no.\ 66)
    \item \texttt{Japan\_Immigrant} (Japan no.\ 301)
  \end{itemize}
  \item \textbf{Base agent (1).} Uses the same general instruction template as participants, but without any profile injection.
  \item \textbf{Moderator (1).} Uses a dedicated system prompt that enforces neutrality, turn-taking, and faithful summarization of the public information pool.
\end{itemize}

\subsection{Prompting and Context Construction}
\paragraph{System prompts.}
An example profile of ours can be seen in Appendix \ref{app:profile_example} and an example profile of baseline like (USA\_native):
\begin{tcolorbox}[
  enhanced,
  breakable,
  colback=white,
  colframe=black!25,
  boxrule=0.6pt,
  arc=1.5mm,
  left=2mm,right=2mm,top=1mm,bottom=1mm
]
\begin{lstlisting}[style=jsonstyle]
{
  "personal_values_Profile": [
    "family is very important in my life",
    "friends are very important in my life",
    "leisure time is not very important in my life",
    "politics is not at all important in my life",
    "work is very important in my life",
    ...
    "I would like to have immigrants or foreign workers as neighbors",
    "I would like to have homosexuals as neighbors",
    "I would like to have people of a different religion as neighbors",
    ...
    "I disagree that on the whole, men make better political leaders than women do",
    "I disagree that a university education is more important for a boy than for a girl",
    ...
    "I have a great deal of confidence in the police",
    "I have a great deal of confidence in the courts",
    "I do not have very much confidence in the government",
    ...
    "I obtain information daily via internet to learn what's going on in this country and the world",
    "I have signed a petition before",
    ...
    "I am a female",
    "I was born in United States",
    "My mother was born in United States",
    "My father was born in United States",
    ...
    "I am married",
    "I have 2 children",
    "The highest educational level that I have attained is upper secondary education",
    "I belong to the Roman Catholic religion",
    "I am 44 years old",
    "I am currently in United States"
  ]
}
\end{lstlisting}
\end{tcolorbox}

\paragraph{Shared public information pool.}
We maintain a public pool containing the accumulated utterances. At the beginning of each speaking round, all agents condition on the latest pool contents.

\subsection{Logging}
For each conference, we log:
\begin{itemize}\small
  \item all utterances in chronological order (including moderator summaries),
  \item all agreement scores at Stage 1 and after each Stage 2 update step,
  \item metadata: topic ID, individual IDs
\end{itemize}

\section{Evaluation Prompt}
\label{app:restoration_eval_prompt}
The prompt in Table \ref{tab:restoration_eval_prompt} is used when employing Gemini 3 Pro and ChatGPT 5.2 Thinking as evaluators.
\begin{center}
\begin{promptbox}{Evaluation Prompt: VAS \& SSS}
\small
\begin{lstlisting}[basicstyle=\ttfamily\footnotesize,columns=fullflexible,keepspaces=true,showstringspaces=false,breaklines=true,breakatwhitespace=false]
# Role
You are an expert Psychologist and Sociologist specializing in profiling and psycholinguistic analysis.

# Task
Evaluate the "Restoration Degree" of a Simulated AI Agent based strictly on its **Target Profile** and **Discourse (Speeches)**.
Your goal is to determine if the agent successfully "becomes" the specific person described in the profile, or if they remain a generic AI assistant.

# Scoring Metrics (Strict 1-00 Scale)

**1. Value Alignment Score (1-100)**
*Definition:* Do the agent's arguments and stance align with the specific values (e.g., WVS answers, political leanings, religious views) in the Profile?
- **1 (Contradiction):** Arguments directly oppose the profile 
- **50 (General Alignment):** Broadly consistent with the stance, but lacks specific reference to the profile's unique value system.
- **100 (Perfect Match):** Arguments explicitly anchor on specific values from the profile.

**2. Style simulation Score (1-100)**
*Definition:* Does the linguistic style (tone, vocabulary, complexity, sentence structure) match the Demographic Background (Age, Education, Class)?
- **1 (Generic AI):** Sounds like a standard AI assistant 
- **100 (Authentic Voice):** Sounds like a real person of that specific age and class. 

# Output Format (JSON)
Please strictly output the result in the following JSON format:

{
  "Agent_ID": "Person [ID]",
  "Scores": {
    "Value Alignment Score": 0,
    "Style simulation Score": 0
  },
  "Reasoning": "Concise analysis (2-3 sentences). Quote a specific sentence from the transcript that either proves or disproves the persona restoration."
}
\end{lstlisting}
\end{promptbox}
\captionsetup{hypcap=false}
\captionof{table}{Evaluation prompt for assessing persona restoration in dynamic discourse.}
\label{tab:restoration_eval_prompt}
\captionsetup{hypcap=true}
\end{center}

\section{Detailed Results in Multi-Agent Debate}
\label{app:debate_vas_sss_tables}

This appendix provides detailed results for the multi-agent debate experiments described in Section~\ref{subsec:nexus_eval} and complements the aggregate results in Table~\ref{tab:pdi_results}. The two debate topics are:\\
\textbf{Accept Newcomers}: Willing to accept newcomers as members of your community.\\
\textbf{Local or Universal}: Artistic creation should serve local values rather than pursue universal expression.\\
We report per-individual \textbf{Value Alignment Score (VAS)}, \textbf{Style Simulation Score (SSS)}, and \textbf{Persona Distinctiveness Index (PDI)} for these two topics. We compare \textbf{ExpertIVS} with \textbf{IndieValue}, \textbf{Simple Concatenation}, and \textbf{Anthology}. VAS and SSS are evaluated by \textit{Gemini 3 Pro} and \textit{ChatGPT 5.2 Thinking} over five independent runs, and the averaged per-individual results are reported in Tables~\ref{tab:accept_gemini_appendix}, \ref{tab:accept_chatgpt_appendix}, \ref{tab:local_gemini_appendix}, and \ref{tab:local_chatgpt_appendix}.

\begin{table}[t]
\centering
\scriptsize
\setlength{\tabcolsep}{2.5pt}
\resizebox{\columnwidth}{!}{%
\begin{tabular}{lcccccccc}
\toprule
& \multicolumn{2}{c}{ExpertIVS} & \multicolumn{2}{c}{IndieValue} & \multicolumn{2}{c}{Simple} & \multicolumn{2}{c}{Anthology} \\
\cmidrule(lr){2-3}\cmidrule(lr){4-5}\cmidrule(lr){6-7}\cmidrule(lr){8-9}
\textbf{Individual} & \textbf{VAS} & \textbf{SSS} & \textbf{VAS} & \textbf{SSS} & \textbf{VAS} & \textbf{SSS} & \textbf{VAS} & \textbf{SSS} \\
\midrule
USA\_Native        & 97.40 & 68.40 & 96.80 & 29.00 & 90.00 & 5.00  & 90.00 & 15.00 \\
USA\_Immigrant     & 94.00 & 42.00 & 61.00 & 30.00 & 90.00 & 15.00 & 85.00 & 25.00 \\
Germany\_Native    & 96.80 & 59.60 & 98.00 & 43.00 & 85.00 & 10.00 & 85.00 & 20.00 \\
Germany\_Immigrant & 96.20 & 50.00 & 87.60 & 18.00 & 95.00 & 5.00  & 95.00 & 15.00 \\
Russia\_Native     & 98.00 & 59.00 & 96.20 & 35.00 & 90.00 & 10.00 & 90.00 & 10.00 \\
Russia\_Immigrant  & 98.00 & 78.00 & 98.00 & 48.00 & 85.00 & 10.00 & 90.00 & 20.00 \\
Japan\_Native      & 96.20 & 67.00 & 96.20 & 32.00 & 80.00 & 15.00 & 95.00 & 15.00 \\
Japan\_Immigrant   & 98.00 & 85.20 & 96.80 & 50.00 & 95.00 & 15.00 & 90.00 & 15.00 \\
\midrule
\textbf{Overall}   & 96.83 & 63.65 & 91.33 & 35.62 & 88.75 & 10.63 & 90.00 & 16.88 \\
\bottomrule
\end{tabular}
}
\caption{Per-individual VAS and SSS on the \textit{Accept Newcomers} debate topic, evaluated by Gemini 3 Pro.}
\label{tab:accept_gemini_appendix}
\end{table}

\begin{table}[t]
\centering
\scriptsize
\setlength{\tabcolsep}{2.5pt}
\resizebox{\columnwidth}{!}{%
\begin{tabular}{lcccccccc}
\toprule
& \multicolumn{2}{c}{ExpertIVS} & \multicolumn{2}{c}{IndieValue} & \multicolumn{2}{c}{Simple} & \multicolumn{2}{c}{Anthology} \\
\cmidrule(lr){2-3}\cmidrule(lr){4-5}\cmidrule(lr){6-7}\cmidrule(lr){8-9}
\textbf{Individual} & \textbf{VAS} & \textbf{SSS} & \textbf{VAS} & \textbf{SSS} & \textbf{VAS} & \textbf{SSS} & \textbf{VAS} & \textbf{SSS} \\
\midrule
USA\_Native        & 86.00 & 34.20 & 74.60 & 31.00 & 86.00 & 34.00 & 88.00 & 62.00 \\
USA\_Immigrant     & 80.00 & 32.40 & 41.20 & 27.20 & 84.00 & 38.00 & 86.00 & 72.00 \\
Germany\_Native    & 90.00 & 40.00 & 85.60 & 28.00 & 72.00 & 38.00 & 76.00 & 78.00 \\
Germany\_Immigrant & 88.00 & 39.60 & 71.20 & 24.80 & 78.00 & 34.00 & 78.00 & 62.00 \\
Russia\_Native     & 88.00 & 36.00 & 63.20 & 25.80 & 87.00 & 58.00 & 92.00 & 62.00 \\
Russia\_Immigrant  & 86.80 & 35.00 & 80.00 & 36.20 & 78.00 & 44.00 & 92.00 & 72.00 \\
Japan\_Native      & 91.40 & 34.00 & 81.20 & 24.40 & 66.00 & 34.00 & 90.00 & 62.00 \\
Japan\_Immigrant   & 88.40 & 41.80 & 72.80 & 30.00 & 86.00 & 32.00 & 90.00 & 62.00 \\
\midrule
\textbf{Overall}   & 87.33 & 36.62 & 71.22 & 28.43 & 79.63 & 39.00 & 86.50 & 66.50 \\
\bottomrule
\end{tabular}
}
\caption{Per-individual VAS and SSS on the \textit{Accept Newcomers} debate topic, evaluated by ChatGPT 5.2 Thinking.}
\label{tab:accept_chatgpt_appendix}
\end{table}

\begin{table}[t]
\centering
\scriptsize
\setlength{\tabcolsep}{2.5pt}
\resizebox{\columnwidth}{!}{%
\begin{tabular}{lcccccccc}
\toprule
& \multicolumn{2}{c}{ExpertIVS} & \multicolumn{2}{c}{IndieValue} & \multicolumn{2}{c}{Simple} & \multicolumn{2}{c}{Anthology} \\
\cmidrule(lr){2-3}\cmidrule(lr){4-5}\cmidrule(lr){6-7}\cmidrule(lr){8-9}
\textbf{Individual} & \textbf{VAS} & \textbf{SSS} & \textbf{VAS} & \textbf{SSS} & \textbf{VAS} & \textbf{SSS} & \textbf{VAS} & \textbf{SSS} \\
\midrule
USA\_Native        & 95.00 & 85.00 & 90.00 & 15.00 & 85.00 & 20.00 & 65.00 & 40.00 \\
USA\_Immigrant     & 95.00 & 15.00 & 90.00 & 30.00 & 35.00 & 20.00 & 95.00 & 40.00 \\
Germany\_Native    & 95.00 & 25.00 & 85.00 & 20.00 & 95.00 & 15.00 & 95.00 & 30.00 \\
Germany\_Immigrant & 85.00 & 25.00 & 95.00 & 15.00 & 75.00 & 15.00 & 90.00 & 35.00 \\
Russia\_Native     & 95.00 & 30.00 & 75.00 & 10.00 & 45.00 & 15.00 & 80.00 & 35.00 \\
Russia\_Immigrant  & 95.00 & 45.00 & 75.00 & 15.00 & 95.00 & 70.00 & 50.00 & 35.00 \\
Japan\_Native      & 92.00 & 80.00 & 35.00 & 5.00  & 75.00 & 5.00  & 95.00 & 50.00 \\
Japan\_Immigrant   & 95.00 & 40.00 & 95.00 & 15.00 & 65.00 & 15.00 & 95.00 & 90.00 \\
\midrule
\textbf{Overall}   & 93.38 & 43.13 & 80.00 & 15.63 & 71.25 & 21.88 & 83.13 & 44.38 \\
\bottomrule
\end{tabular}
}
\caption{Per-individual VAS and SSS on the \textit{Local or Universal} debate topic, evaluated by Gemini 3 Pro.}
\label{tab:local_gemini_appendix}
\end{table}

\begin{table}[t]
\centering
\scriptsize
\setlength{\tabcolsep}{2.5pt}
\resizebox{\columnwidth}{!}{%
\begin{tabular}{lcccccccc}
\toprule
& \multicolumn{2}{c}{ExpertIVS} & \multicolumn{2}{c}{IndieValue} & \multicolumn{2}{c}{Simple} & \multicolumn{2}{c}{Anthology} \\
\cmidrule(lr){2-3}\cmidrule(lr){4-5}\cmidrule(lr){6-7}\cmidrule(lr){8-9}
\textbf{Individual} & \textbf{VAS} & \textbf{SSS} & \textbf{VAS} & \textbf{SSS} & \textbf{VAS} & \textbf{SSS} & \textbf{VAS} & \textbf{SSS} \\
\midrule
USA\_Native        & 76.00 & 63.00 & 72.00 & 38.00 & 58.00 & 22.00 & 56.00 & 72.00 \\
USA\_Immigrant     & 86.00 & 83.00 & 72.00 & 36.00 & 56.00 & 34.00 & 92.00 & 34.00 \\
Germany\_Native    & 78.00 & 34.00 & 68.00 & 22.00 & 78.00 & 34.00 & 78.00 & 38.00 \\
Germany\_Immigrant & 78.00 & 44.00 & 68.00 & 22.00 & 58.00 & 34.00 & 72.00 & 34.00 \\
Russia\_Native     & 73.00 & 38.00 & 58.00 & 35.00 & 56.00 & 24.00 & 62.00 & 38.00 \\
Russia\_Immigrant  & 78.00 & 61.00 & 78.00 & 52.00 & 62.00 & 34.00 & 58.00 & 33.00 \\
Japan\_Native      & 62.00 & 44.00 & 68.00 & 22.00 & 54.00 & 18.00 & 78.00 & 72.00 \\
Japan\_Immigrant   & 72.00 & 44.00 & 72.00 & 34.00 & 62.00 & 38.00 & 74.00 & 56.00 \\
\midrule
\textbf{Overall}   & 75.38 & 51.38 & 69.50 & 32.63 & 60.50 & 29.75 & 71.25 & 47.12 \\
\bottomrule
\end{tabular}
}
\caption{Per-individual VAS and SSS on the \textit{Local or Universal} debate topic, evaluated by ChatGPT 5.2 Thinking.}
\label{tab:local_chatgpt_appendix}
\end{table}

\section{Human Evaluation Details}
\label{app:human_eval}

We additionally conducted human evaluation for the “Accept Newcomers” setting. We recruited three evaluators with diverse backgrounds, including a government office staff member, a sociology master’s student, and an IT industry employee. Each evaluator received USD 70 in shopping vouchers as compensation. To avoid bias, our method and the baseline (IndieValueCatalog) were anonymized as Method A and Method B during evaluation. The human evaluation protocol was consistent with the LLM-based evaluation rubric, and the detailed scores of the three evaluators are reported in Table~\ref{tab:human_eval_scores}.

\begin{table}[t]
\centering
\small
\resizebox{\columnwidth}{!}{%
\begin{tabular}{lcccc}
\toprule
\textbf{Individual} & \textbf{Ours\_VAS} & \textbf{Ours\_SSS} & \textbf{IndieValue\_VAS} & \textbf{IndieValue\_SSS} \\
\midrule
USA-native        & 89.67 & 56.67 & 80.67 & 43.33 \\
USA-immigrant     & 82.33 & 53.33 & 81.67 & 46.67 \\
Germany-native    & 92.67 & 51.67 & 84.00 & 41.67 \\
Germany-immigrant & 93.00 & 54.00 & 84.67 & 40.67 \\
Russia-native     & 86.67 & 50.00 & 81.00 & 44.33 \\
Russia-immigrant  & 86.67 & 54.33 & 83.67 & 51.67 \\
Japan-native      & 93.67 & 57.67 & 83.00 & 51.67 \\
Japan-immigrant   & 95.67 & 55.67 & 79.00 & 45.00 \\
\midrule
\textbf{Average}  & \textbf{90.04} & \textbf{54.17} & \textbf{82.21} & \textbf{45.63} \\
\bottomrule
\end{tabular}
}
\caption{Human evaluation scores on the \textit{Accept Newcomers} setting, averaged over three evaluators.}
\label{tab:human_eval_scores}
\end{table}

\end{document}